%% file: main.tex
\documentclass[10pt,twocolumn,letterpaper]{article}

\usepackage{cvpr}              % To produce the CAMERA-READY version

\usepackage{times}
\usepackage{epsfig}
\usepackage{graphicx}
\usepackage{amsmath}
\usepackage{amssymb}
\usepackage{booktabs}
\usepackage{flushend}
\usepackage[dvipsnames]{xcolor}
\definecolor{cvprblue}{rgb}{0.21,0.49,0.74}
\usepackage[pagebackref,breaklinks,colorlinks,citecolor=cvprblue]{hyperref}

\def\paperID{343} % *** Enter the Paper ID here
\def\confName{3DV\xspace}
\def\confYear{2024\xspace}

\input{commands}
\input{variables}

\input{vincents_commands}

\addeditor{vincent}{VL}{0.0, 0.5, 0.0}
\addeditor{gilles}{GP}{0.0, 0.5, 0.9}
\addeditor{renaud}{RM}{0.95, 0.55, 0.15}
\addeditor{alex}{AB}{0.6, 0.4, 1.0}
\addeditor{corentin}{CS}{1.0, 0.0, 1.0}
\showeditsfalse

\title{BEVContrast: Self-Supervision in BEV Space for Automotive Lidar Point Clouds}

\author{
Corentin Sautier$^{1,2}$ \quad
Gilles Puy$^{2}$ \quad
Alexandre Boulch$^{2}$ \quad
Renaud Marlet$^{1,2}$ \quad
Vincent Lepetit$^{1}$ \quad
\and
\and
\large
\hspace{-3mm} \textsuperscript{1}LIGM, Ecole des Ponts, Univ Gustave Eiffel, CNRS, Marne-la-Vall\'ee, France
\hspace{1mm} \textsuperscript{2}Valeo.ai, Paris, France 
}

\begin{document}

\maketitle
\input{sections/abstract}
\input{sections/introduction}
\input{sections/related}
\input{sections/method}
\input{sections/experiments}
\input{sections/conclusion}
{
    \small
    \bibliographystyle{ieeenat_fullname}
    \bibliography{main}
}
\input{sections/supmat}

% \pagebreak
% {
%     \small
%     \bibliographystyle{ieeenat_fullname}
%     \bibliography{main}
% }

\end{document}

%% file: commands.tex
\usepackage[utf8]{inputenc} % allow utf-8 input
\usepackage[T1]{fontenc}    % use 8-bit T1 fonts
\usepackage{amsfonts}       % blackboard math symbols
\usepackage{nicefrac}       % compact symbols for 1/2, etc.
\usepackage{microtype}      % microtypography

\usepackage{pifont}% http://ctan.org/pkg/pifont
\usepackage{colortbl}
\usepackage{multirow}
\usepackage{adjustbox}

\usepackage{minitoc}

\def\ours{BEVContrast\xspace} 
\newcommand{\smallparagraph}[1]{\medskip\noindent\textbf{#1}~}
\newcommand{\Rbb}{{\mathbb{R}}}

\definecolor{MyGreen}{RGB}{0, 180, 0}
\definecolor{MyRed}{RGB}{180, 0, 0}
\definecolor{MyBlue}{RGB}{30, 0, 180}

\newcommand{\tplus}[1]{\textcolor{MyGreen}{#1}}

\newcommand{\std}[1]{\footnotesize $\pm${#1}}

\newcommand{\rtwo}[1]{\cellcolor{MyGreen!10}{#1}}
\newcommand{\rone}[1]{\cellcolor{MyGreen!30}{#1}}

\newcolumntype{R}[2]{%
    >{\adjustbox{angle=#1,lap=1.3\width-(#2)}\bgroup}%
    l%
    <{\egroup}%
}

\definecolor{skcar}{RGB}{110,160,245}
\definecolor{skotherv}{RGB}{0, 0, 255}
\definecolor{skdrivable}{RGB}{255, 0,255}
\definecolor{sksidewalk}{RGB}{175, 0,175}
\definecolor{skterrain}{RGB}{190, 210, 130}
\definecolor{skvegetation}{RGB}{0, 175, 0}
\definecolor{skfence}{RGB}{225, 180, 135}
\definecolor{sktrunk}{RGB}{180, 120, 13}
\definecolor{skparking}{RGB}{225, 180, 225}

\newcommand{\greyrule}[1]{\arrayrulecolor{black!20} #1\arrayrulecolor{black}}

%% file: variables.tex
\newcommand{\pc}{\mathcal{P}}
\newcommand{\pcone}{\mathcal{P}}
\newcommand{\pctwo}{\mathcal{P}'}
\newcommand{\bevone}{\mathcal{B}}
\newcommand{\bevtwo}{\mathcal{B}'}

\newcommand{\bevstride}{b}%{b_s}

%% file: vincents_commands.tex
\usepackage{dsfont}
\usepackage{etoolbox}

\newif\ifshowedits

\newcommand{\addeditor}[3]{%
  \definecolor{#1color}{rgb}{#3}
  \expandafter\newcommand\csname #1\endcsname[1]{%
  \ifshowedits
    {\color{#1color} ##1}%
  \else
    {##1}%
  \fi
  }%
  \expandafter\newcommand\csname #1rmk\endcsname[1]{%
  \ifshowedits
    {\color{#1color} {\bf [#2: ##1]}}
  \fi
  }%
  \expandafter\newcommand\csname #1rpl\endcsname[2]{%
  \ifshowedits
    {\color{#1color} ##1 \sout{##2}}
  \else
    {##1}
  \fi
  }%
}

\newcommand{\createtextvar}[1]{
  \expandafter\newcommand\csname #1\endcsname{%
  {\text{#1}}
}%
}
\newcommand{\mycomment}[1]{}

%% file: sections/abstract.tex
\begin{abstract}
We present a surprisingly simple and efficient method for self-supervision of 3D backbone on automotive Lidar point clouds. We design a contrastive loss between features of Lidar scans captured in the same scene. Several such approaches have been proposed in the literature from PointConstrast~\cite{pointcontrast}, which uses a contrast at the level of points, to the state-of-the-art TARL~\cite{TARL}, which uses a contrast at the level of segments, roughly corresponding to objects. While the former enjoys a great simplicity of implementation, it is surpassed by the latter, which however requires a costly pre-processing. In \ours, we define our contrast at the level of 2D cells in the Bird’s Eye View plane. Resulting cell-level representations offer a good trade-off between the point-level representations exploited in PointContrast and segment-level representations exploited in TARL: we retain the simplicity of PointContrast (cell representations are cheap to compute) while surpassing the performance of TARL in downstream semantic segmentation.
The code is available at \href{https://github.com/valeoai/BEVContrast}{github.com/valeoai/BEVContrast}
\end{abstract}

%% file: sections/introduction.tex
% =========================================================

\section{Introduction}

\begin{figure*}
\small
\def\svgwidth{\linewidth}
\input{figures/methods_latex.tex}
\caption{
\textbf{Comparison of several self-supervised contrastive methods with \ours.}
SegContrast~\cite{segcontrast} and TARL~\cite{TARL} learn powerful representations by contrasting features at the level of segments obtained thanks to a careful and costly pre-processing. PointContrast~\cite{pointcontrast} is simpler as it requires no pre-processing and directly contrasts representations at the point level, but experiments show that these point-level representations are not as powerful as segment-level representations. \ours preserves the simplicity of PointContrast by extracting cell-level representations in BEV. These representations are cheap to compute, require no pre-processing and perform better than  SegContrast and TARL for downstream semantic segmentation. ($\oplus$ and $\ominus$ represent positive and negative contrastive losses respectively; $t_1$ and $t_2$ stand for two different time steps.)
}
\label{fig:methods}
\end{figure*}
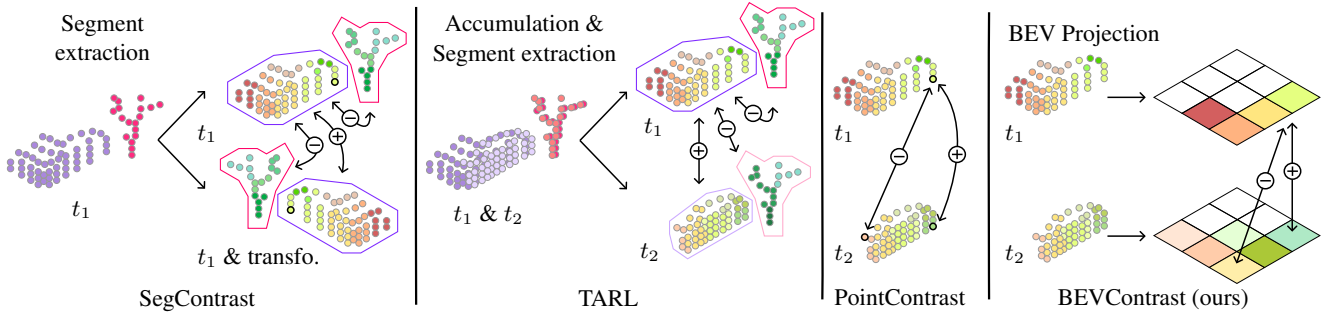

Lidar point cloud processing has recently received a great deal of attention, in particular because of its applicability in the field of autonomous driving. However, annotating a single Lidar point cloud for training a deep architecture can take multiple hours~\cite{semantickitti}, while deep networks in safety-critical autonomous cars will require training on many scans of very diverse scenes, multiplying the need for such annotations. An attractive solution is self-supervised learning, which has been shown to reduce the need for labeled data by pre-training a model on a pretext task requiring no label~\cite{dino,byol,once,slidr}. Urban scenes are especially suitable for this kind of pre-training, as the cost of obtaining data is only a fraction of the cost of annotating it.

Self-supervised methods based on a contrastive loss are among the best performing approaches. They are constructed on the following principle: deep representations extracted from two views (i.e., two scans) of a scene with sufficient overlap should be identical. PointContrast~\cite{pointcontrast} uses this principle at the point level, while SegContrast~\cite{segcontrast} and TARL~\cite{TARL} use it at the level of segments that roughly correspond to objects. After identifying pairs of corresponding points or segments, the contrastive loss forces (i)~the representations in each pair to be as similar as possible while (ii)~keeping representations in two different pairs as dissimilar as possible. An illustration of the application of this contrastive principle at the point and segment levels is available in Figure~\ref{fig:methods}.

The advantage of considering point-level representations as in PointContrast is the simplicity of the method: these representations are readily available at the output of the 3D backbone. Experiments nevertheless show that point-level representations do not generalize as well as segment-level representations to semantic downstream tasks~\cite{TARL,segcontrast}. Unfortunately, the latter require
a complex segmentation step as pre-processing, with the addition of several hyperparameters to tune. For example, TARL requires temporal aggregation of multiple scans, careful road plane detection to remove ground points, and point clustering on the remaining points to create point cloud segments to contrast.

In this work, we propose to contrast features at the level of 2D cells of a grid on the Bird's Eye View~(BEV) plane. The feature of a cell is here defined as the average of the features of the points projecting into that cell.
We call the resulting method \ours. It is motivated by the fact that objects in urban scenes are naturally well separated in the BEV plane. Therefore, locally averaging point features in this plane permits us to obtain cell-level features, which are a good trade-off between point-level representations, and segment-level representations.  
The high-level principle of \ours is illustrated in Figure~\ref{fig:methods}. It retains the simplicity of PointContrast (the projection in BEV and local average pooling in each BEV cell being cheap to compute) while experiments show that it competes with the best self-supervised methods such as TARL, despite the fact that we do not treat dynamic objects explicitly.

Our contributions are the following. First, we propose a novel self-supervised method for Lidar point clouds that retains the simplicity of PointContrast while surpassing all concurrent self-supervised methods for downstream semantic segmentation\renaud{, including non-contrastive methods too, such as ALSO~\cite{also}}. Second, we show that projecting and pooling features in bird's eye views is surprisingly more effective than pooling features over segments extracted with complex methods. Finally, as \ours works in bird's eye views, it can be applied out of the box to pre-train backbones for 3D object detection such as SECOND~\cite{second} or PVRCNN~\cite{pvrcnn}, with which we achieve competing results compared to the state-of-the-art.

%% file: figures/methods_latex.tex
%% Creator: Inkscape 1.2 (dc2aedaf03, 2022-05-15), www.inkscape.org
%% Accompanies image file 'methods_latex.pdf' (pdf, eps, ps)
\begingroup%
  \makeatletter%
  \providecommand\color[2][]{%
    \errmessage{(Inkscape) Color is used for the text in Inkscape, but the package 'color.sty' is not loaded}%
    \renewcommand\color[2][]{}%
  }%
  \providecommand\transparent[1]{%
    \errmessage{(Inkscape) Transparency is used (non-zero) for the text in Inkscape, but the package 'transparent.sty' is not loaded}%
    \renewcommand\transparent[1]{}%
  }%
  \providecommand\rotatebox[2]{#2}%
  \newcommand*\fsize{\dimexpr\f@size pt\relax}%
  \newcommand*\lineheight[1]{\fontsize{\fsize}{#1\fsize}\selectfont}%
  \ifx\svgwidth\undefined%
    \setlength{\unitlength}{1749.04835138bp}%
    \ifx\svgscale\undefined%
      \relax%
    \else%
      \setlength{\unitlength}{\unitlength * \real{\svgscale}}%
    \fi%
  \else%
    \setlength{\unitlength}{\svgwidth}%
  \fi%
  \global\let\svgwidth\undefined%
  \global\let\svgscale\undefined%
  \makeatother%
  \begin{picture}(1,0.23495884)%
    \lineheight{1}%
    \setlength\tabcolsep{0pt}%
    \put(0,0){\includegraphics[width=\unitlength,page=1]{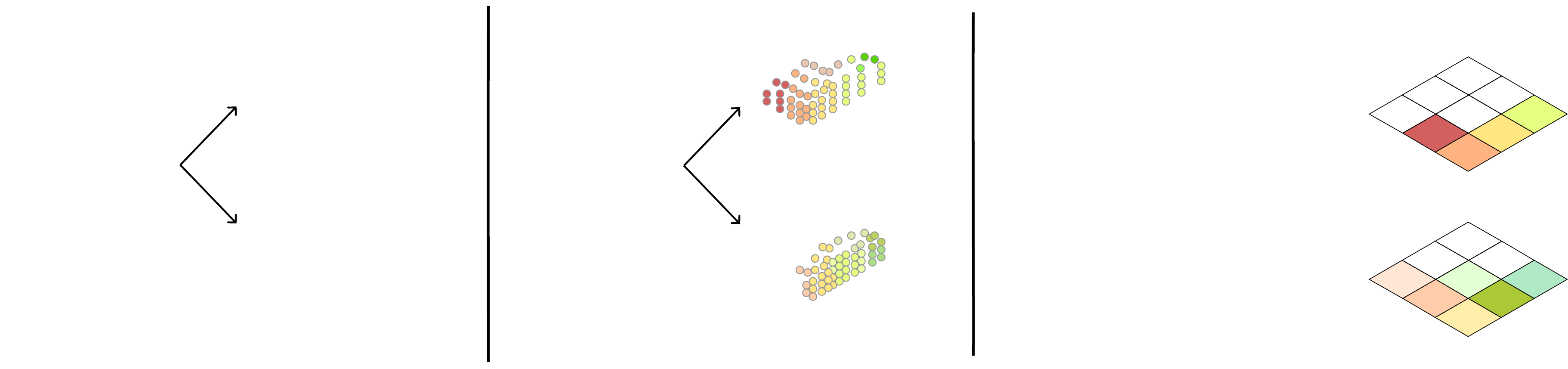}}%
    \put(0.485,0.135){\makebox(0,0)[lt]{\lineheight{1.25}\smash{\begin{tabular}[t]{l}$t_1$\end{tabular}}}}%
    \put(0.482,0.0395){\makebox(0,0)[lt]{\lineheight{1.25}\smash{\begin{tabular}[t]{l}$t_2$\end{tabular}}}}%
    \put(0,0){\includegraphics[width=\unitlength,page=2]{figures/resources/methods_latex.pdf}}%
    \put(0.145,0.13){\makebox(0,0)[lt]{\lineheight{1.25}\smash{\begin{tabular}[t]{l}$t_1$\end{tabular}}}}%
    \put(0,0){\includegraphics[width=\unitlength,page=3]{figures/resources/methods_latex.pdf}}%
    \put(0.34,0.06774038){\makebox(0,0)[lt]{\lineheight{1.25}\smash{\begin{tabular}[t]{l}$t_1$ \& $t_2$\end{tabular}}}}%
    \put(0,0){\includegraphics[width=\unitlength,page=4]{figures/resources/methods_latex.pdf}}%
    \put(0.10074571,0.003455){\makebox(0,0)[lt]{\lineheight{1.25}\smash{\begin{tabular}[t]{l}SegContrast\end{tabular}}}}%
    \put(0.435,0.00324984){\makebox(0,0)[lt]{\lineheight{1.25}\smash{\begin{tabular}[t]{c}TARL\end{tabular}}}}%
    \put(0,0){\includegraphics[width=\unitlength,page=5]{figures/resources/methods_latex.pdf}}%
    \put(0.76,0.13){\makebox(0,0)[lt]{\lineheight{1.25}\smash{\begin{tabular}[t]{l}$t_1$\end{tabular}}}}%
    \put(0.76,0.038){\makebox(0,0)[lt]{\lineheight{1.25}\smash{\begin{tabular}[t]{l}$t_2$\end{tabular}}}}%
    \put(0,0){\includegraphics[width=\unitlength,page=6]{figures/resources/methods_latex.pdf}}%
    \put(0.63,0.13){\makebox(0,0)[lt]{\lineheight{1.25}\smash{\begin{tabular}[t]{l}$t_1$\end{tabular}}}}%
    \put(0.63,0.038){\makebox(0,0)[lt]{\lineheight{1.25}\smash{\begin{tabular}[t]{l}$t_2$\end{tabular}}}}%
    \put(0,0){\includegraphics[width=\unitlength,page=7]{figures/resources/methods_latex.pdf}}%
    \put(0.63,0.00391151){\makebox(0,0)[lt]{\lineheight{1.25}\smash{\begin{tabular}[t]{c}PointContrast\end{tabular}}}}%
    \put(0.8,0.00391151){\makebox(0,0)[lt]{\lineheight{1.25}\smash{\begin{tabular}[t]{l}BEVContrast (ours)\end{tabular}}}}%
    \put(0,0){\includegraphics[width=\unitlength,page=8]{figures/resources/methods_latex.pdf}}%
    \put(0.145,0.035){\makebox(0,0)[lt]{\lineheight{1.25}\smash{\begin{tabular}[t]{l}$t_1$ \& transfo.\end{tabular}}}}%
    \put(0.0728945,0.21297255){\makebox(0,0)[t]{\lineheight{1.25}\smash{\begin{tabular}[t]{c}Segment\\extraction\end{tabular}}}}%
    \put(0,0){\includegraphics[width=\unitlength,page=9]{figures/resources/methods_latex.pdf}}%
    \put(0.394,0.21289792){\makebox(0,0)[t]{\lineheight{1.25}\smash{\begin{tabular}[t]{c}Accumulation \&\\Segment extraction\\\end{tabular}}}}%
    \put(0.82,0.205){\makebox(0,0)[t]{\lineheight{1.25}\smash{\begin{tabular}[t]{c}BEV Projection\\\end{tabular}}}}%
    \put(0.04939107,0.072){\makebox(0,0)[lt]{\lineheight{1.25}\smash{\begin{tabular}[t]{l}$t_1$\end{tabular}}}}%
  \end{picture}%
\endgroup%

%% file: sections/related.tex
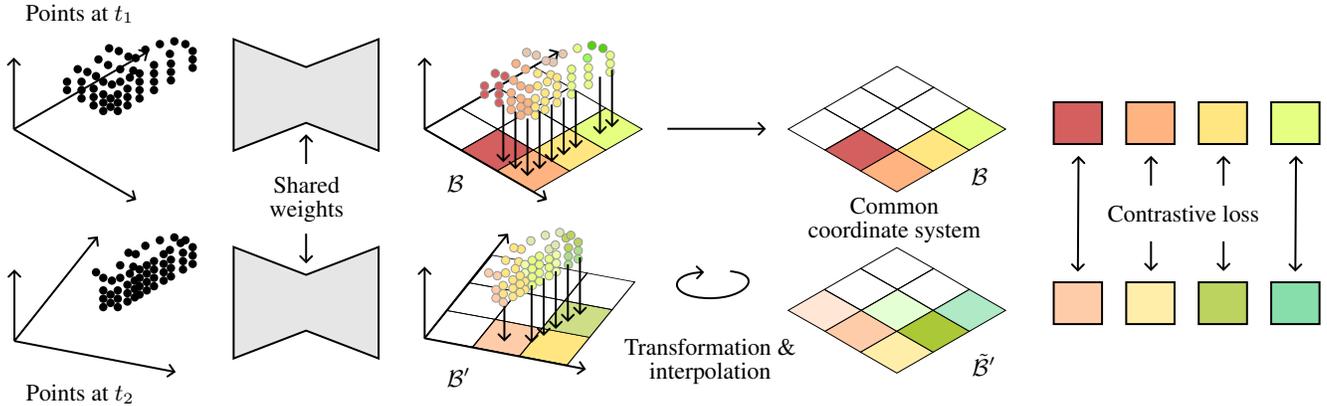
\begin{figure*}[t]
\small
\def\svgwidth{\textwidth}
\input{figures/overview_latex.tex}
\caption{\textbf{Overview of BEVContrast.} Two point cloud views of the same scenes are encoded and projected on the BEV plane. One BEV is aligned to the other's coordinate frame by affine transformation and interpolation, and a loss constrasts the features in both BEV views.
}
\label{fig:overview}
\end{figure*}

\section{Related work}
\label{sec:related}

Self-supervision consists in learning representations by creating a pretext task from the data itself without the need of any annotations. 
The goal is to obtain good representations that generalize well to many downstream tasks where limited annotations are available. In this section, we provide an overview of self-supervised training methods working on images, point clouds, or both modalities.

\smallparagraph{Self-supervision on images.}
Early self-supervised methods defined pretext tasks such as predicting by which amount an image was rotated~\cite{rotnet} or solving a jigsaw puzzle~\cite{jigsaw}. Today, the most successful class of self-supervised methods either leverage discriminative tasks~\cite{simclr,moco,byol,obow,dino,swav} in which a network is trained to extract representations that are invariant to augmentations or a masked image modeling task~\cite{mae,beit} in which the network is trained to reconstruct image parts hidden from the network input.

\smallparagraph{Self-supervision on point clouds.} 
The advances on self-supervision on point clouds followed closely the improvements made on images. Early self-supervised methods were using pretext tasks such as predicting transformation applied on the point cloud or reconstructing parts of the point cloud~\cite{Poursaeed2020SelfSupervisedLO,sauder2019self}. These methods were applied on dense scans of single objects.

The advent of discriminative based methods permitted the development of efficient self-supervised methods working on entire indoor or outdoor scenes.  PointContrast~\cite{pointcontrast}, DepthContrast~\cite{depthcontrast} and STRL~\cite{strl} were the among the first such methods.
Concerning methods targetting specifically automotive Lidar data, ProposalContrast~\cite{proposalcontrast}, SegContrast~\cite{segcontrast} and TARL~\cite{TARL} construct a discriminative tasks by pooling features at the level of segments that roughly correspond to objects.
The three methods first filter out ground points, e.g., using RANSAC~\cite{ransac} for SegConstant or Patchwork~\cite{patchwork} for TARL, and then extract segments either as a spherical region~\cite{proposalcontrast} or using HDBSCAN~\cite{hdbscan}.
In addition, TARL leverages the temporal dimension and aggregates multiple scans to generate temporally-consistent segments.
The temporal axis is also used in STSSL~\cite{stssl}, which combines point-level contrastive learning within each segment and segment-level contrastive learning between frames.
In contrast, our method shows that simply projecting and pooling features in BEV is surprisingly more effective than pooling over such segments.

Finally, reconstruction-based methods are also successful for self-supervision on point clouds. Masked image modeling tasks have also been adapted to point clouds data. Some methods reconstruct points coordinates using the Chamfer distance~\cite{pointm2ae,gdmae,pang2022masked,hess2023masked}, other predict voxel occupancy~\cite{voxelmae}, or predict point-patch token from a codebook ~\cite{pointbert,fu2023boosting}.
% Recently, ALSO~\cite{also} proposed to use unsupervised surface reconstruction as a pretext task to train 3D backbone on automotive Lidar point clouds and reached high performance on downstream semantic segmentation and object detection. 
\corentin{Recently, ALSO~\cite{also} proposes to use unsupervised surface reconstruction as a pretext task to train 3D backbones on automotive Lidar point clouds. Using the knowledge of occupancy before and after \renaud{an observed 3D point along a Lidar line of sight,} % an intersection of a laser ray 
it learns to construct an implicit \renaud{occupancy} function and, incidentally, good point features. \renaud{Although it does not exploit contrast, it} % without contrast, and 
reaches high performance on downstream semantic segmentation and object detection.}

\smallparagraph{Multi-modal self-supervision.} Another line of work leverages synchronized and calibrated cameras and Lidar to pre-train a 3D backbone~\cite{simipu,ppkt,slidr,st_slidr}. The underlying idea is to find pairs of corresponding points and pixels and ensure that the associated point and pixel representations are as close as possible.
We note that the image backbone has to be pre-trained on an external dataset, e.g., ImageNet~\cite{imagenet}, for \cite{ppkt,slidr,st_slidr}.
While these methods can obtain impressive results, the use of another modality as well as an external image dataset makes them not comparable to our method.

%% file: figures/overview_latex.tex
%% Creator: Inkscape 1.2 (dc2aedaf03, 2022-05-15), www.inkscape.org
%% Accompanies image file 'overview_latex.pdf' (pdf, eps, ps)
\begingroup%
  \makeatletter%
  \providecommand\color[2][]{%
    \errmessage{(Inkscape) Color is used for the text in Inkscape, but the package 'color.sty' is not loaded}%
    \renewcommand\color[2][]{}%
  }%
  \providecommand\transparent[1]{%
    \errmessage{(Inkscape) Transparency is used (non-zero) for the text in Inkscape, but the package 'transparent.sty' is not loaded}%
    \renewcommand\transparent[1]{}%
  }%
  \providecommand\rotatebox[2]{#2}%
  \newcommand*\fsize{\dimexpr\f@size pt\relax}%
  \newcommand*\lineheight[1]{\fontsize{\fsize}{#1\fsize}\selectfont}%
  \ifx\svgwidth\undefined%
    \setlength{\unitlength}{1334.48315237bp}%
    \ifx\svgscale\undefined%
      \relax%
    \else%
      \setlength{\unitlength}{\unitlength * \real{\svgscale}}%
    \fi%
  \else%
    \setlength{\unitlength}{\svgwidth}%
  \fi%
  \global\let\svgwidth\undefined%
  \global\let\svgscale\undefined%
  \makeatother%
  \begin{picture}(1,0.31924755)%
    \lineheight{1}%
    \setlength\tabcolsep{0pt}%
    \put(0,0){\includegraphics[width=\unitlength,page=1]{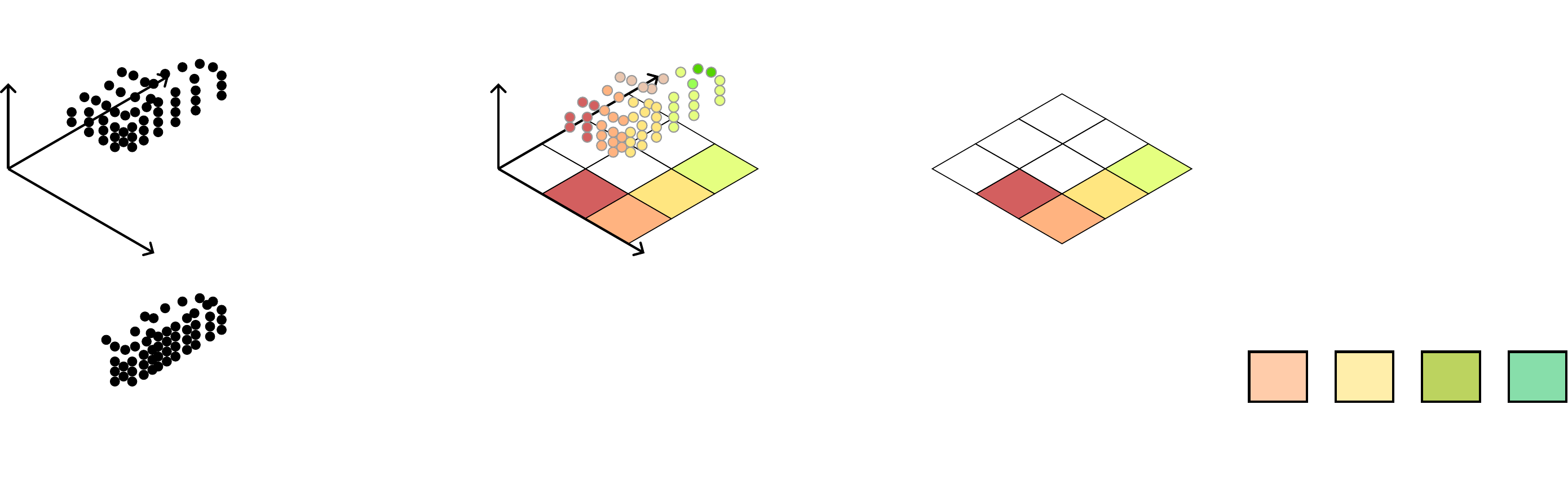}}%
    \put(0.01419452,0.29452341){\makebox(0,0)[lt]{\lineheight{1.}\smash{\begin{tabular}[t]{l}Points at $t_1$\end{tabular}}}}%
    \put(0.01444333,0.0045283){\makebox(0,0)[lt]{\lineheight{1.}\smash{\begin{tabular}[t]{l}Points at $t_2$\end{tabular}}}}%
    \put(0,0){\includegraphics[width=\unitlength,page=2]{figures/resources/overview_latex.pdf}}%
    \put(0.2,0.163){\makebox(0,0)[lt]{\lineheight{1.}\smash{\begin{tabular}[t]{c}Shared\\weights\end{tabular}}}}%
    \put(0.47,0.04){\makebox(0,0)[lt]{\lineheight{1.}\smash{\begin{tabular}[t]{c}Transformation \&\\interpolation\end{tabular}}}}%
    \put(0.61,0.147){\makebox(0,0)[lt]{\lineheight{1.}\smash{\begin{tabular}[t]{c}Common\\coordinate system\end{tabular}}}}%
    \put(0.838,0.141){\makebox(0,0)[lt]{\lineheight{1.}\smash{\begin{tabular}[t]{c}Contrastive loss\end{tabular}}}}%
    \put(0.335,0.16383974){\makebox(0,0)[lt]{\lineheight{1.}\smash{\begin{tabular}[t]{l}$\bevone$\end{tabular}}}}%
    \put(0.73397302,0.16904841){\makebox(0,0)[lt]{\lineheight{1.}\smash{\begin{tabular}[t]{l}$\bevone$\end{tabular}}}}%
    \put(0,0){\includegraphics[width=\unitlength,page=3]{figures/resources/overview_latex.pdf}}%
    \put(0.73479466,0.023552){\makebox(0,0)[lt]{\lineheight{1.}\smash{\begin{tabular}[t]{l}$\tilde{\mathcal{B}}'$\end{tabular}}}}%
    \put(0,0){\includegraphics[width=\unitlength,page=4]{figures/resources/overview_latex.pdf}}%
    \put(0.335,0.01514907){\makebox(0,0)[lt]{\lineheight{1.}\smash{\begin{tabular}[t]{l}$\bevtwo$\end{tabular}}}}%
  \end{picture}%
\endgroup%

%% file: sections/method.tex
\section{Method}
\label{sec:method}

Our method (\ours) is designed to pre-train a 3D backbone $f_\theta(\cdot)$ from Lidar scans without any annotations, where $\theta$ represents the learnable parameters. We describe below the method for 3D segmentation backbones, i.e., when $f_\theta(\cdot)$ takes as input a point cloud $\pc$ and outputs
a $D$-dimensional feature vector for each point. We denote the feature vector for the $i^{\rm th}$ point in $\pc$ by $f_\theta(\pc)_i \in \Rbb^D$. 
However, the method is also usable for 3D object detection with minor modifications, as described in Section~\ref{sec:odmethod}.

% ============================================
\subsection{Pre-training Data}

\ours requires pairs of partially overlapping Lidar scans $(\pcone, \pctwo)$ captured in a same scene but from different viewpoints. Typically, these point clouds will be captured at different instants separated by a few seconds~$\Delta_{\rm time}$. Each point of $\pcone$ or $\pctwo$ is described by a 4-dimensional vector containing its 3D Cartesian $xyz$-coordinates and the measured return intensity~(reflectance).

We also need the 3D rotation matrix $R \in \Rbb^{3 \times 3}$ and 3D translation vector $t \in \Rbb^3$ that register $\pctwo$ to $\pcone$. $R$ and $t$ can be computed from the absolute or relative poses of the Lidar, which are usually available in autonomous driving datasets.

% ============================================
\subsection{BEV pooling}

As discussed in the introduction, PointContrast enjoys a great simplicity of implementation by working directly on point-level representations. However, these learned representations do not rival with segment-level representations such as those in SegContrast and TARL~\cite{segcontrast,TARL}. Yet, the higher performance of SegContrast and TARL, compared to PointContrast, comes at the cost of an expensive pre-processing that requires extra hyperparameters to tune.

In this work, instead of pooling point representations over
expensive-to-get segments, we propose to project features on the BEV plane and locally pool features over 2D cells of a grid defined on this plane. We show in the next section that this simple and fast-to-compute pooling mechanism is key to reach state-of-the-art results.

Our pooling mechanism is constructed on the fact that most of the objects in a Lidar scan are naturally well separated in the BEV plane. We thus propose to project the points in $\pc$ in the BEV plane of height $z=0$, divide this plane in regular cells of size $\bevstride\times\bevstride$, and average the representations in $f_\theta(\pc)$ of points falling in a same BEV cell. Thanks to this process, and for a reasonable cell size~$\bevstride$, each BEV cell feature describes an object or parts of an object with little ``contamination'' from neighboring objects. These cell-level features are thus a good proxy for object-level representations.
Mathematically, we denote these BEV representations $\bevone = g(f_\theta(\pcone), \pcone) \in \Rbb^{M \times M \times D}$ and $\bevtwo = g(f_\theta(\pctwo), \pctwo) \in \Rbb^{M \times M \times D}$, where $M \times M$ is the number of BEV cells, and $g(\cdot, \cdot)$ represents the projection into BEV and average pooling step. Cells with no point projected into them are filled with null features.

% ============================================
\subsection{Contrastive loss}

We construct our self-supervised contrastive loss directly on the BEV representations.
The first step is to align the BEV representations $\bevone$ and  $\bevtwo$ to put cells in correspondences. In practice, we exploit the information contained in $R$ and $t$ to compute a 2D affine transformation which we use to register $\bevtwo$ onto $\bevone$ thanks to a bilinear interpolation. We denote by $\tilde{\mathcal{B}}'$ the feature map $\bevtwo$ transformed into the same BEV grid as $\bevone$.
We provide the details about this affine transformation in the supplementary material, and a study of alternative registration methods in Table~\ref{table:ablation_registration}.

We pre-train $f_\theta(\cdot)$ using a discriminative task at the level of BEV cells. Let us denote $\tilde{\mathcal{B}}_{l}' \in \mathrm{R}^D$ and $\bevone_{m} \in \mathrm{R}^D$ the $D$-dimensional representations extracted from $\tilde{\mathcal{B}}'$ and $\bevone$ in the $l^{\rm th}$ and $m^{\rm th}$ BEV cells, respectively. Our loss enforces BEV representations $\bevone_{l}$ and $\tilde{\mathcal{B}}_{l}'$ falling into the same cell $l$ to be identical, while keeping BEV representations $\bevone_{m}$ and $\tilde{\mathcal{B}}_{l}'$ falling in two different cells $l$ and $m$ as different as possible. Concretely, we sample at random a set $\mathcal{N}$ of non-empty BEV cells and minimize
\begin{align}
    \mathcal{L}(\theta) = 
    - \sum_{l \in \mathcal{N}} \log \left[  
    \frac
    {\exp\left({\tilde{\mathcal{B}}_{l}'} \cdot \bevone_{l} / \tau\right)}
    {\sum_{m \in \mathcal{N}} \exp\left({\tilde{\mathcal{B}}_{l}' \cdot \bevone_{m}} / \tau \right)}
    \right],
\end{align}
where $\tau>0$ is a temperature hyperparameter and $\cdot$ denotes the scalar product in $\mathbb{R}^D$. Note that the BEV cell features are $\ell_2$-normalized before computing the above loss.

\subsection{Adaptation to 3D detection}
\label{sec:odmethod}

Note that \ours can be applied out of the box to pre-train popular object detection backbones such as those used in SECOND \cite{second} or PVRCNN \cite{pvrcnn}. Indeed, these backbones output, by design, features in a BEV plane grid, hence, we do not need to apply any projection.
In this case, the BEV pooling and choice of $\bevstride$ is omitted as it is taken care of by the backbone, while the affine transformation parameters are obtained as above.

%% file: sections/experiments.tex
\section{Experiments}
\label{sec:experiments}

In this section, we compare \ours to state-of-the-art methods for self-supervision. We compare the performance of these methods using two downstream tasks: semantic segmentation in Section~\ref{sec:semseg} and object detection in Section~\ref{sec:od}. Then, we study the sensitivity of \ours to its hyperparameters in Section~\ref{sec:sensitivity}. We conclude in Section~\ref{sec:limitations} with a discussion about the advantages and disadvantages of \ours and the concurrent methods.

\input{tables/results.tex}

\input{figures/visus_semseg}

\subsection{Datasets}
\label{sec:datasets}

\smallparagraph{nuScenes (NS)~\cite{nuscenes}}~contains 700 and 150 scenes for training and validation, acquired in Boston and Singapore with a 32-beam Lidar. For a fair comparison with the related work, we pre-train our network using 600 training scenes and exploit the remaining 100 training scenes to tune all hyperparameters. The performance of all methods are compared on the official validation set of nuScenes, which thus plays the role of the test set. Note that nuScenes sequences contain unannotated scans that could be used for self-supervision: scans are captured at 20\,Hz but in fact only annotated at 2\,Hz. \corentin{However, to maintain comparability with previous work, even during pre-training,}
% We experiment here in the same setting as previous work: for self-supervised pre-training, 
we only use the scans that are annotated, although we ignore the annotations.
The pre-trained backbones are then fine-tuned on different subsets of the training scenes. One can refer to \cite{slidr,also,st_slidr} for more details about this protocol, which we follow exactly.

\smallparagraph{SemanticKITTI (SK)~\cite{semantickitti}}~contains 10 training sequences and 1 validation sequence captured with a 64-beam Lidar in various environments in the vicinity of Karlsruhe in Germany. We exploit the full training set for pre-training and the partial training sets defined in \cite{segcontrast} for the downstream tasks.

\smallparagraph{SemanticPOSS (SP)~\cite{semanticposs}}~is composed of 6 sequences captured on the campus of the Peking University with a 40-beam Lidar and annotated with a subset of the labels classes of SemanticKITTI. 
We use this dataset in Section~\ref{sec:limitations} to test the capacity of self-supervised pre-trained backbones in generalizing to different Lidars.
We fine-tune the pre-trained backbones using the partial training sets used in~\cite{also} and evaluate the performance on the official validation set.

\smallparagraph{KITTI 3D Detection (K)~\cite{kitti3d}}~consists in 7.5k Lidar frames split into a training set and a validation set. The annotations are 3D bounding boxes around cars, bicyclists and pedestrians. Unlike the previously described datasets, KITTI3D does not contain any information that allows one to reconstruct a temporal sequence of scans. This prevents us to pre-train any backbone on this dataset, but we can use it for downstream object detection.

\subsection{Semantic segmentation}
\label{sec:semseg}

In this section, we pre-train 3D backbones for semantic segmentation on either nuScenes or SemanticKITTI, and evaluate the quality of the pre-training by downstream fine-tuning on subsets of nuScenes or SemanticKITTI. Each of these subsets contains either $0.1 \%$, $1 \%$, $10 \%$, $50 \%$ or $100 \%$ of the annotated training data. All self-supervised pre-trained backbones are fine-tuned using the protocol of \cite{also}. For methods not already evaluated in \cite{also}, we obtained the pre-trained backbones directly from the authors and run the fine-tuning experiments ourselves.

% \subsubsection{Experimental setting}
\label{sec:exp_setting_sem_seg}

\smallparagraph{3D backbones.}~We follow the common practice in the literature (see, e.g., \cite{also}) and pre-train a ResUNet18 on SemanticKITTI and a ResUNet34 on nuScenes.

\smallparagraph{\ours pre-training protocol.}
We pre-train the backbones with \ours for $50$ epochs with a batch size of~$10$, split across two A100-40GB GPUs for SemanticKITTI and a batch size of $24$ on a single A100 for nuScenes. \renaud{The pre-training uses} % We recall that that we use 
the full training set of SemanticKITTI, and 600 training scenes out of the 700 available for nuScenes (see \cite{also} for details).
We use AdamW \cite{adamw} with a learning rate of $0.001$, a weight decay of $0.001$, $\mathcal{N} = 4096$ samples in the loss, a temperature hyperparameter $\tau$ of $0.07$ as in~\cite{pointcontrast, moco} and a cosine annealing scheduler. 
The hyperparameters $\bevstride$ and $\Delta_{\rm time}$ have been set to $30$\,cm and $1$\,s on nuScenes and $20$\,cm and $0.7$\,s on SemanticKITTI. We limit the size $M$ of the BEV to $512$ for SemanticKITTI and $256$ for nuScenes.

\smallparagraph{Fine-tuning protocol.}~After pre-training, the backbones are fine-tuned for semantic segmentation on different subsets of SemanticKITTI or nuScenes.
We use the fine-tuning hyperparameters and splits defined in~\cite{also}. Specifically, we use a batch size of~8 for nuScenes, and 2 for  SemanticKITTI.
We compute the per-point final score by setting each point's prediction as the voxel prediction it fall in. 
For each pre-trained model and each subset, we perform 5 independent fine-tunings and report the average and standard deviation of the mIoU over these runs.

\smallparagraph{Other details.}~All experiments on SemanticKITTI (pre-training and fine-tuning) in \cite{segcontrast,TARL,also} are done by filtering out the few points labelled as ``ignore.'' In our method, we use all points during pre-training, as no labels should be used during this stage, but filter out these points during fine-tuning. Hence, all fine-tuning experiments are done in the same setting as the above methods.

\smallparagraph{Results.} We compare in Table~\ref{table:results} the performance of several pre-training methods. We observe that
\ours is the state-of-the-art method on all datasets and all splits in the case where the backbones are pre-trained and fine-tuned on the same datasets. \corentin{Surprisingly, per-class results (\renaud{Table}~\ref{tab:per_class_sk} \renaud{in the appendix}) do not show a significant difference between static and mobile objects, even though TARL's segmentation should be robust to them. We repeat the observation that contrastive methods based on segment pooling are better performing than point-level \renaud{contrast}.} We also present qualitative results obtained with different methods in Figure~\ref{fig:visu_semseg}.

% =============================================================
\subsection{Detection}
\label{sec:od}
\input{tables/detection}

In this section, we evaluate the interest of using our method to pre-train two widely-used object detectors: SECOND~\cite{second} and PVRCNN~\cite{pvrcnn}. We compare \ours to methods which already pre-train these object detectors and for which results are available in the literature. Note that these object detectors are \emph{not} pre-trained in TARL \cite{TARL} and, as it is not straightforward to adapt TARL for this backbone, we do not report any comparison with TARL for object detection.

\smallparagraph{Backbone.} We pre-train the backbone which is shared by the object detectors SECOND and PVRCNN. This backbone is made of a 3D sparse encoder, which outputs a BEV representation of the input point cloud, and is followed by a 2D encoder, which refines the BEV representation.

\smallparagraph{BEVContrast pre-training protocol.} We pre-train this backbone with \ours for $50$ epochs with a batch size of $10$ on a single A100-40GB GPUs on the 600 training scenes of nuScenes reserved for pre-training. We use AdamW \cite{adamw} with a learning rate of $0.001$, a weight decay of $0.001$, a temperature $\tau$ of $0.07$ and a cosine annealing scheduler. As we target downstream fine-tuning on KITTI3D, we use the OpenPCDet voxel size tuned for this dataset: $5\,{\rm cm}$ on the $x$ and $y$ axes, $10\,{\rm cm}$ on the $z$ axis. We limit the range to a square of side $102.4\,{\rm m}$ centered on the ego-car during pretraining, as done in OpenPCDet when training on nuScenes.
We do not use any extra pooling after the 2D encoder: the parameter $\bevstride$ is fixed by the BEV resolution and the convolution strides chosen in SECOND~\cite{second} and PVRCNN~\cite{pvrcnn}.

\smallparagraph{Fine-tuning protocol.} After pre-training, the detection head of SECOND or PVRCNN is appended to the pre-trained backbone and the whole network is fine-tuned on KITTI3D. We use the OpenPCDet~\cite{openpcdet} implementation of these object detectors and the default OpenPCDet training parameters. As in \cite{also}, we perform 3 independent fine-tunings and report the best mAP on the validation set of KITTI3D.

\smallparagraph{Results.} We present the results obtained with \ours and concurrent methods in Table~\ref{table:detection}. First, we notice that pre-training with \ours always leads to better results than without any pre-training at all. Second, when comparing methods that are also pre-trained on nuScenes, we notice that \ours performs better than GCC-3D and competes with ALSO, while being significantly simpler to implement. Finally, when considering all methods and all pre-training datasets, \ours is in the top 3 of all methods. \corentin{Please note that this experiment \renaud{involves a single run for each method, rather than averaging the performance across multiple runs.} % is not averaged across multiple runs, and that 
The error margin % for each of those methods 
\renaud{may thus} be significant\renaud{, especially as the variance is generally higher for object detection than for semantic segmentation.} Nevertheless, contrastive methods\renaud{, inluding ours,} % and ours specifically 
seem \renaud{slightly} less effective on object detection compared to semantic segmentation. This could arise from the nature of detection, \renaud{which is a more} geometrical task, compared to segmentation, \renaud{which is} a per-point classification task, as contrastive learning tends not to encourage the learning of geometric object positioning, \renaud{contrary to} reconstruction\renaud{-based} methods such as ALSO.}

% =============================================================
\subsection{Sensitivity to hyperparameters}
\label{sec:sensitivity}

In this section, we study the sensitivity of \ours to its hyperparameters. We use the following experimental protocol. All models are pre-trained for 20 epochs on either nuScenes or SemanticKITTI with a batch size of~2. The models are then fine-tuned on their pre-training datasets exactly as described in Section~\ref{sec:exp_setting_sem_seg}, except for SemanticKITTI where the batch size is set to~$4$. We use the splits with $10\%$ of annotations for fine-tuning.

\subsubsection{Sensitivity to $\Delta_{\rm time}$ and $\bevstride$}
\label{sec:ablation}

We start by studying the sensitivity of our method to the choice of the hyperparameters $\Delta_{\rm time}$, the time difference between the acquisitions of $\pcone$ and $\pctwo$, and $\bevstride$, the size of the BEV cells.

We present the results obtained when pre-training and fine-tuning on nuScenes in Table~\ref{table:ablation_nuscenes}. We do not notice dramatic drop of performance for any of tested pairs $(\Delta_{\rm time}, b)$.
The best results are obtained at $\bevstride=30\,{\rm cm}$ and $\Delta_{\rm time} = 1.0\,{\rm s}$, which are the parameters we used to produce the results in Table~\ref{table:results}.

The results obtained when pre-training and fine-tuning on SemanticKITTI are presented in Table~\ref{table:ablation_sk}. The set of tested parameters is smaller than for nuScenes, but we still notice that the results are relatively stable at $\bevstride=20\ {\rm cm}$ for all tested $\Delta_{\rm time}$, with a variation of at most $1.0$ points of mIoU. The results presented in Table~\ref{table:results} were obtained by using $\bevstride=20\ {\rm cm}$ and $\Delta_{\rm time} = 0.7\ {\rm s}$.

\input{tables/ablation_sk}

\subsubsection{Choice of the scans to contrast}
\label{sec:ablation_dx}
\input{tables/ablation_dx}

Many different strategies exist to select overlapping point clouds $\pcone$ and $\pctwo$ in \ours. We study two simple alternatives in this section: selecting scans acquired $\Delta_{\rm time}$ seconds apart, or selecting scans acquired after a displacement of (at least) $\Delta_{\rm dist}$ meters of the ego-car.

We conduct this study on SemanticKITTI with $\bevstride=20$~cm and report the results in Table~\ref{table:ablation_dx}. We notice that the performance is rather robust whether we select scans based on the time of acquisition or the displacement of the ego-car. On SemanticKITTI, we achieve the best results with $\Delta_{\rm time} = 0.7$~s. For experiments on the other datasets, we privileged a selection based on $\Delta_{\rm time}$, as it also easier to implement as sequences are acquired with a fixed frame rate.

\begin{figure*}
    \small
    \centering
    \includegraphics[width=\linewidth]{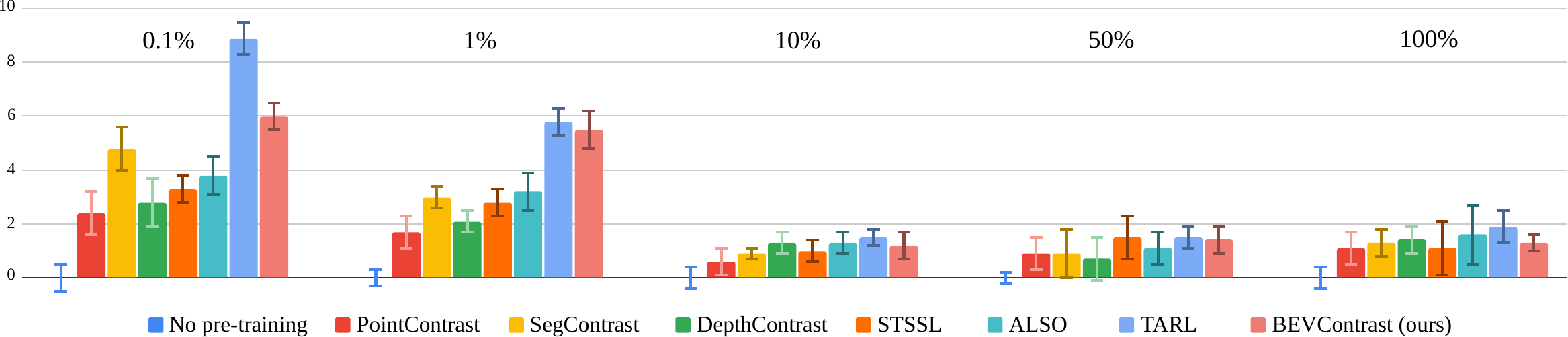}
    \caption{\textbf{Semantic segmentation fine-tuning results on SemanticPOSS} after pre-training on SemanticKITTI. Different subsets from $0.1\%$ to $100\%$ of the training set of SemanticPOSS are used for fine-tuning. We report the relative improvement compare to \textit{no pre-training} in mIoU\%, average and standard deviation over 5 different fine-tunings. The individual fine-tuning results are in the supplementary material.
    }
    \label{fig:sk_poss_transfer}
\end{figure*}

\subsubsection{Study of different registration methods}
\label{sec:ablation_registration}

\ours requires aligned BEV representations for pre-training the 3D backbone $f_\theta(\cdot)$. In this section, we study different strategies to obtain these aligned representations:
\begin{enumerate}
    \item The first option, which we denote by ``\emph{3D},'' consists in registering $\pctwo$ after computing $f_\theta(\pctwo)$ but before projection in BEV. We compute $g(f_\theta(\pctwo), r(\pctwo))$ where $r(\cdot)$ applies the rigid 3D transformation given by $R$ and $t$.
    \item The second option consists in aligning the representations after projection in BEV, as proposed in Section~\ref{sec:method}, but using a nearest neighbor interpolation instead of bilinear interpolation. We denote this option by ``\emph{2D NN.}''
    \item The third option consists in the alignment described in  Section~\ref{sec:method}, i.e., using a bilinear interpolation. We denote this option by ``\emph{2D Bi.}'' 
\end{enumerate}
Note that ``\emph{2D NN}'' and ``\emph{2D Bi.}'' can be used when pre-training object detection backbones, but not ``\emph{3D}'' as these backbones often output directly a BEV representation where height information is lost.

\input{tables/ablation_registration}

The results are presented in Table~\ref{table:ablation_registration}. We notice that 2D registration with bilinear interpolation leads to much better results than with nearest neighbor interpolation. Interestingly as well, 3D registration does not perform as well as 2D registration with bilinear interpolation. We hypothesize that approximate registration of the BEV representations with ``\emph{2D Bi.}'' acts as a form of regularizer which avoids overfitting and leads to 3D representations that generalize better to the downstream task.

% =============================================================
\subsection{Limitation and Discussion}
\label{sec:limitations}

\smallparagraph{Transfer between datasets.}
We study the edge case of generalization to data acquired with a Lidar different from the one used for the pre-training data.
Let us mention nevertheless that, in the context of self-supervision, this setting is mostly relevant in the rare case where one is unable to collect sufficient data for self-supervision in the setting of interest. 
We consider SemanticKITTI as the pre-training dataset and fine-tune the pre-trained backbone on SemanticPOSS. The results are presented in Figure~\ref{fig:sk_poss_transfer}.
We observe that \ours achieves competitive results and is ranked second or third on every annotated portion of the training set of SemanticPOSS. Furthermore, it is in the error margin with the SOTA on all but the 0.1\% split, as when using more annotated scans, the ranking of the methods becomes much less clear as the variability of the results is large.

\smallparagraph{Sequential data.}
STSSL~\cite{stssl}, TARL~\cite{TARL} and \ours, and more generally all methods that use two or more frames, require the datasets to be provided as sequential data.
It is the case for most datasets~\cite{nuscenes,semantickitti,waymo,once}, but not for KITTI detection benchmark, where frame timestamps are not provided.
This prevents us from to pre-train on KITTI.
In comparison, other methods such as ALSO~\cite{also} and Voxel-MAE~\cite{voxelmae} process one frame at a time, allowing them to pre-train on any dataset, but making them unable to exploit the redundancy of sequential acquisition.

\smallparagraph{Pre-processing.} An advantage of \ours is the absence of pre-processing before pre-training while, e.g., the pre-processing in TARL costs about 45\,s/frame on a single core of a Xeon 4114 CPU on scans of SemanticKITTI. Furthermore, the hyperparameters for road detection and segment extraction in \cite{TARL,segcontrast} are numerous (minimum cluster size, condition to merge different cluster, etc.)\ and difficult to tune when changing the type of Lidar, e.g., from the 64-beam Lidar in SemanticKITTI to the 32-beam Lidar in nuScenes or 40-beam Lidar in SemanticPOSS. Instead, for \ours, the sensitivity study shows that choosing $b \approx 20$\,cm and $\Delta_{\rm time} \approx 1$\,s is not far from optimal on both nuScenes and SemanticKITTI.

\smallparagraph{False negatives and object segmentation.}
In an ideal setting, one would like to contrast features at the level of actual objects, but objects are unknown in the annotation-less self-supervision scenario, and have somehow to be approximated. In practice, BEV cells in BEVContrast do not always align well on objects: large objects are over-split into several cells, while a single cell may sometimes contain more than one actual object. Undersegmentation and oversegmentation occur similarly with the segmentation methods of SegContrast and TARL. Fortunately, contrastive approaches are somewhat robust to such point grouping errors, yielding nevertheless good features. 
% In this context, segmentation seems more appealing, as resulting segments tend to look more like actual objects. However, 
Our finding is that, surprisingly, our cell-level point grouping, despite its simplicity, apparent coarseness, and absence of filtering of dynamic objects, is actually more effective than complex segmentation methods that try to discover objects. 
It even outperforms TARL in downstream fine-tuning on both nuScenes and SemanticKITTI in all tested scenarios, although TARL uses time information across successive scans to produce better segments.

%% file: tables/results.tex
\begin{table*}[!ht]
% \small
\setlength{\tabcolsep}{8pt}
\centering

\begin{tabular}{@{}ll|cc|cc|cc|cc|cc@{}}

\toprule
Dataset & Method  & \multicolumn{2}{c}{0.1\%} & \multicolumn{2}{c}{1\%} & \multicolumn{2}{c}{10\%} & \multicolumn{2}{c}{50\%} & \multicolumn{2}{c}{100\%}\\

% NUSCENES PRE-param DOWN-train
\midrule
\multirow{5}{*}{nuScenes} & No pre-training & 21.6 & \std{0.5} & 35.0 & \std{0.3} & 57.3 & \std{0.4} & 69.0 & \std{0.2} & 71.2 & \std{0.2} \\
& PointContrast\dag~\cite{pointcontrast} & \rone{27.1} & \std{0.5} & 37.0 & \std{0.5} & \rtwo{58.9} & \std{0.2} & 69.4 & \std{0.3} & 71.1 & \std{0.2}\\ % PC from ALSO/SLidR
& DepthContrast\dag\cite{depthcontrast} & 21.7 & \std{0.3} & 34.6 & \std{0.5} & 57.4 & \std{0.5} & 69.2 & \std{0.3} & 71.2 & \std{0.2} \\
& ALSO~\cite{also} & 26.2 & \std{0.5} & \rtwo{37.4} & \std{0.3} & \rone{59.0} & \std{0.4} & \rtwo{69.8} & \std{0.2} & \rtwo{71.8} & \std{0.2}\\
& \textit{\ours (ours)} & \rtwo{26.6} & \std{0.5} & \rone{37.9} & \std{0.4} & \rone{59.0} & \std{0.6} & \rone{70.5} & \std{0.2} & \rone{72.2} & \std{0.1} \\

% SEMANTICKITTI
\midrule
\multirow{8}{*}{SemanticKITTI}& No pre-training & 30.0 & \std{0.2} & 46.2 & \std{0.6}  & 57.6 & \std{0.9}  & 61.8 & \std{0.4} & 62.7 & \std{0.3} \\
&PointContrast\ddag~\cite{pointcontrast}	& 32.4 & \std{0.5} & 47.9 & \std{0.5} & 59.7 & \std{0.5} & 62.7 & \std{0.3} & 63.4 & \std{0.4}\\ % PC from SegContrast/TARL
&SegContrast~\cite{segcontrast} & 32.3 & \std{0.3} & 48.9  & \std{0.3} & 58.7  & \std{0.5} & 62.1 & \std{0.4} & 62.3 & \std{0.4}   \\
&DepthContrast\dag~\cite{depthcontrast}	& 32.5 & \std{0.4} & 49.0 & \std{0.4} & 60.3 & \std{0.5} & \rtwo{62.9} & \std{0.5} & \rtwo{63.9} & \std{0.4}\\
& STSSL~\cite{stssl} & 32.0 & \std{0.4} & 49.4 & \std{1.1} & 60.0 & \std{0.6} & \rtwo{62.9} & \std{0.7} & 63.3 & \std{0.3} \\
& ALSO~\cite{also} & 35.0  &  \std{0.1} & 50.0 & \std{0.4} & 60.5  & \std{0.1} & \rone{63.4} & \std{0.5} & 63.6 & \std{0.5}\\
& TARL~\cite{TARL} & \rtwo{37.9} & \std{0.4} & \rtwo{52.5} & \std{0.5} & \rtwo{61.2} & \std{0.3} & \rone{63.4} & \std{0.2} & 63.7 & \std{0.3}\\
& \textit{\ours  (ours)} & \rone{39.7} & \std{0.9} & \rone{53.8} & \std{1.0} & \rone{61.4} & \std{0.4} & \rone{63.4} & \std{0.6} & \rone{64.1} & \std{0.4} \\

\bottomrule
\\[-3.5mm]
\multicolumn{12}{c}{\dag\,as reimplemented by ALSO~\cite{also} \hspace{35pt} \ddag\,as reimplemented by SegContrast~\cite{segcontrast}}\\
\end{tabular}
\vspace*{-3mm}
\caption{\textbf{Semantic segmentation fine-tuning results on nuScenes and SemanticKITTI} using different subsets of the corresponding training set \renaud{for fune-tuning} (from $0.1\%$ to $100\%$). The 3D backbones are pre-trained and fine-tuned on the same datasets. \corentin{A single pre-training is used for each line, and details about pre-training data are given in \renaud{Section}~\ref{sec:datasets}.} We report the average and standard deviation of the mIoU\% over 5 different fine-tunings. Individual classwise fine-tuning results are in the supplementary material.}
\label{table:results}
    
\end{table*}

%% file: figures/visus_semseg.tex
\begin{figure*}
    \small
    \centering
    \setlength{\tabcolsep}{3pt}
    \begin{tabular}{@{}cccc@{}}
        \includegraphics[width=0.24\linewidth]{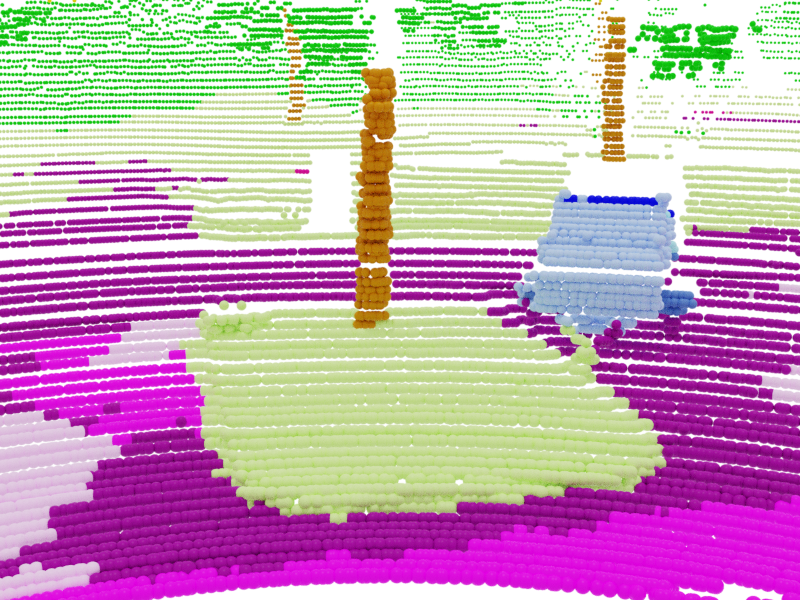} &
        \includegraphics[width=0.24\linewidth]{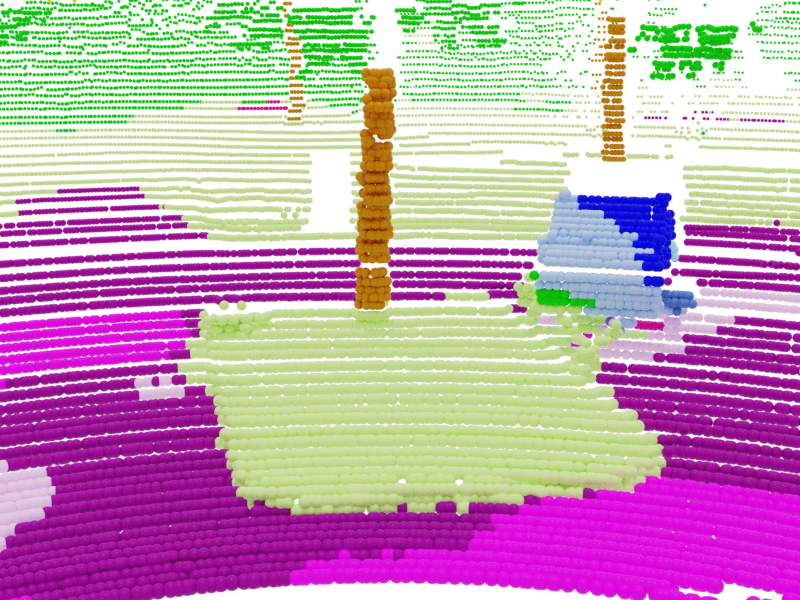} &
        \includegraphics[width=0.24\linewidth]{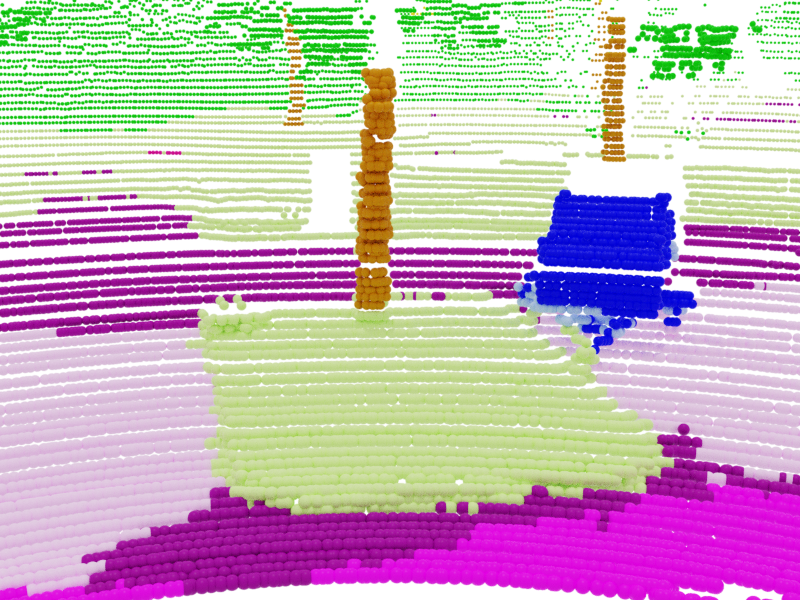} &
        \includegraphics[width=0.24\linewidth]{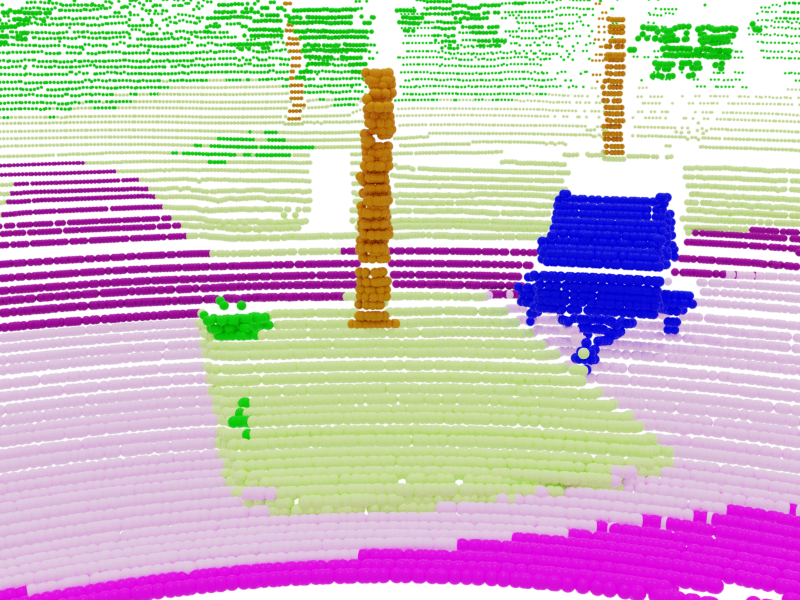} \\
        \includegraphics[width=0.24\linewidth]{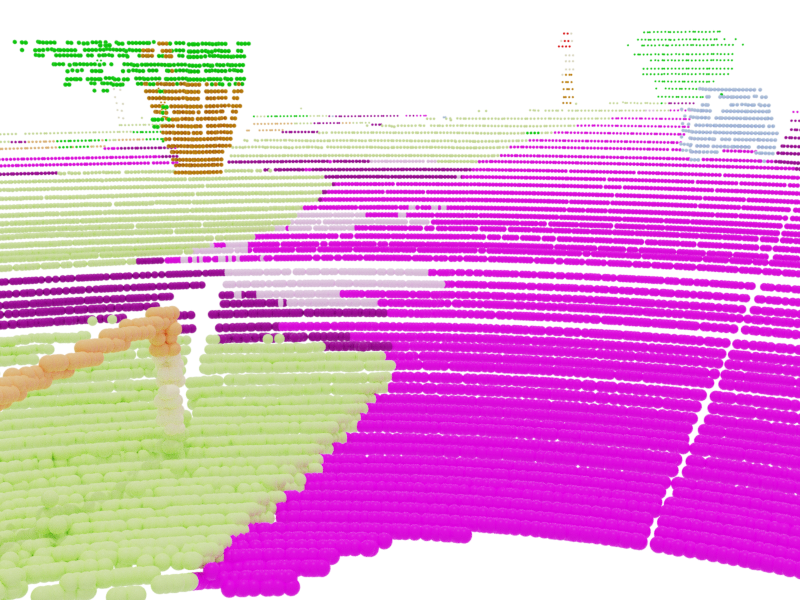} &
        \includegraphics[width=0.24\linewidth]{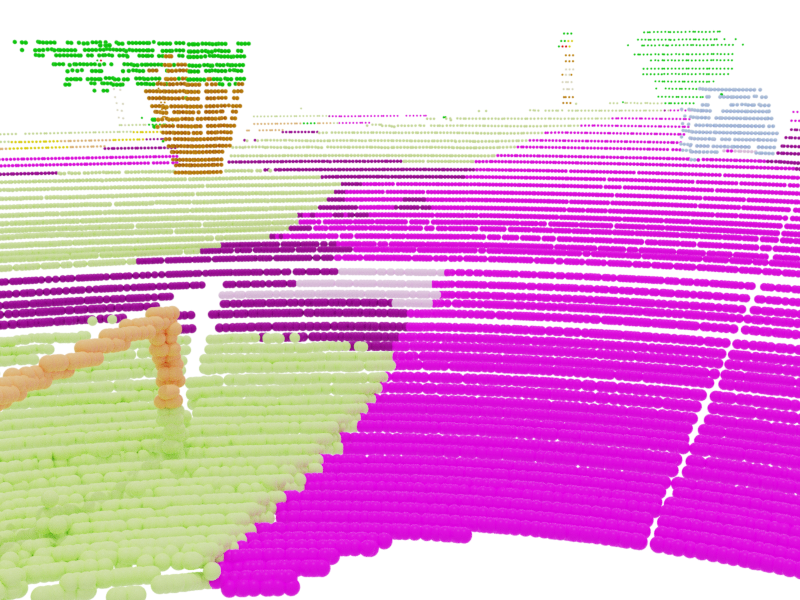} &
        \includegraphics[width=0.24\linewidth]{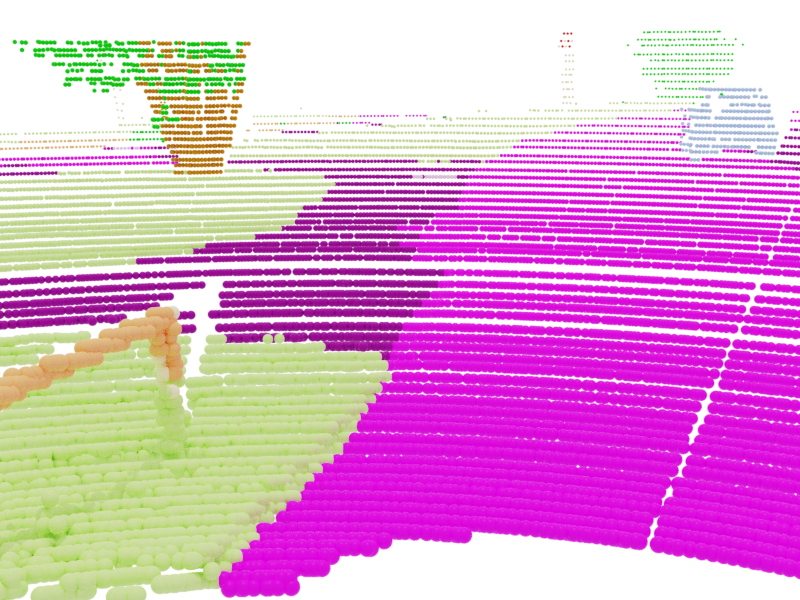} &
        \includegraphics[width=0.24\linewidth]{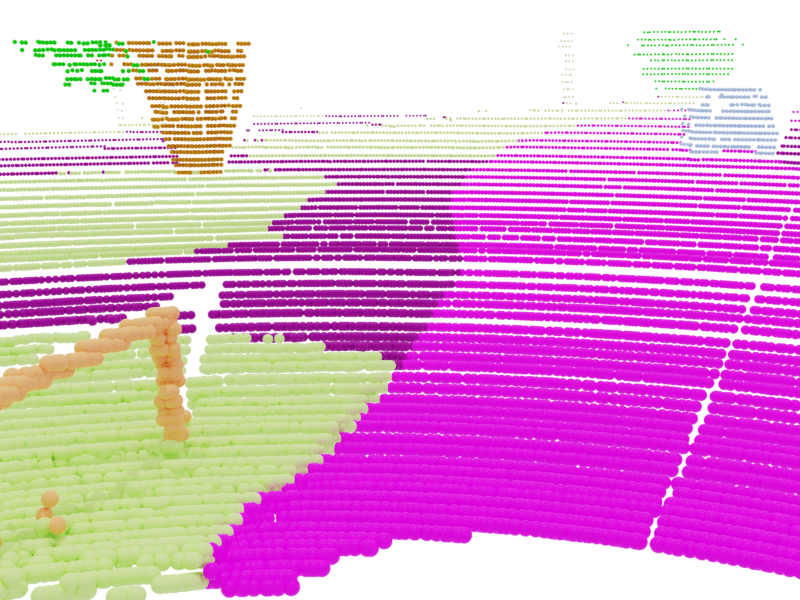} \\
        ALSO & TARL & \ours & Ground truth\\
    \end{tabular}
    \caption{\textbf{Semantic segmentation visualizations on SemanticKITTI} after pre-training on the full training dataset and fine-tuning using 1\% of the annotated data.
    \colorbox{skcar}{\color{white}car\vphantom{$X_p$}},
    \colorbox{skotherv}{\color{white}other veh.\vphantom{$X_p$}},
    \colorbox{sksidewalk}{\color{white}sidewalk\vphantom{$X_p$}},
    \colorbox{skdrivable}{\color{white}drivable surf.\vphantom{$X_p$}},
    \colorbox{skparking}{\color{white}parking\vphantom{$X_p$}},
    \colorbox{skterrain}{\color{white}terrain\vphantom{$X_p$}},
    \colorbox{skvegetation}{\color{white}vegetation\vphantom{$X_p$}},
    \colorbox{sktrunk}{\color{white}trunk\vphantom{$X_p$}},
    \colorbox{skfence}{\color{white}fence\vphantom{$X_p$}}
}
    \label{fig:visu_semseg}
\end{figure*}

%% file: tables/detection.tex
\begin{table}[!t]%[!ht]
\small
\setlength{\tabcolsep}{4.5pt}

\centering

\begin{tabular}{@{}l|c|ccc|cc@{}}
\toprule
 Method & Data. & Cars & Ped. & Cycl. & mAP & Diff.\\

\midrule
\multicolumn{7}{@{}l}{SECOND - $R_{40}$ metric}\\
 No pre-training        & -  & 81.5 & 48.8 & 65.7 & 65.4 &              \\
ALSO~\cite{also}        & NS & \rone{81.8} & \rone{54.2} & \rtwo{68.2} & \rone{68.1} & \tplus{+2.7} \\
\textit{\ours (ours)}   & NS & \rtwo{81.4} & \rtwo{51.9} & \rone{69.3} & \rtwo{67.6} & \tplus{+2.2}\\

\midrule
\multicolumn{7}{@{}l}{SECOND - $R_{11}$ metric}\\

No pre-training             & -  & 78.6 & 53.0 & 67.2 & 66.3 & \\
Voxel-MAE~\cite{voxelmae}   & K  & \rone{78.9} & \rtwo{53.1} & \rtwo{68.1} & 66.7 & \tplus{+0.4} \\
ALSO~\cite{also}            & NS & \rtwo{78.7} & \rone{55.2} & \rtwo{68.1} & \rone{67.3} & \tplus{+1.0} \\
\textit{\ours (ours)}       & NS & 78.3 & 52.8 & \rone{69.3} & \rtwo{66.8} & \tplus{+0.5}\\

\midrule
\multicolumn{7}{@{}l}{PV-RCNN - $R_{40}$ metric}\\

No pre-training                     & -  & 84.5 & 57.1 & 70.1 & 70.6 &  \\
 STRL~\cite{strl}                   & K  & 84.7 & 57.8 & 71.9 & 71.5 & \tplus{+0.9} \\
GCC-3D~\cite{gcc3d}                 & NS & -     & -     & -  & 70.8 & \tplus{+0.2} \\
GCC-3D~\cite{gcc3d}                 & W  & -     & -     & -  & 71.3 & \tplus{+0.7} \\
PointCont.~\cite{pointcontrast}     & W  & 84.2 & 57.7 & 72.7 & 71.6 & \tplus{+1.0} \\
Prop.Cont.~\cite{proposalcontrast}  & W  & 84.7 & \rone{60.4} & 73.7 & \rone{72.9} & \tplus{+2.3} \\
ALSO~\cite{also}                    & NS & \rone{84.9} & \rtwo{57.8} & \rone{75.0} & \rtwo{72.5} & \tplus{+1.9} \\
\textit{\ours (ours)}               & NS & \rtwo{84.8} & 57.3 & \rtwo{74.2} & 72.1 & \tplus{+1.5} \\

\midrule
\multicolumn{7}{@{}l}{PV-RCNN - $R_{11}$ metric}\\

No pre-training             & -  & \rtwo{83.6} & 57.9 & 70.5 & 70.7 & \\
Voxel-MAE~\cite{voxelmae}   & K  & \rone{83.8} & \rone{59.4} & 72.0 & \rtwo{71.7} & \tplus{+1.0} \\
ALSO~\cite{also}            & NS & \rone{83.8} & \rtwo{58.5} & \rone{74.4} & \rone{72.2} & \tplus{+1.5} \\
\textit{\ours (ours)}       & NS & \rtwo{83.6} & 58.0 & \rtwo{73.1} & 71.6 & \tplus{+0.9} \\

\bottomrule

\end{tabular}
\caption{\textbf{Detection results on KITTI3D~\cite{kitti3d}}, validation set, moderate difficulty. We report AP (\%)
and the dataset used for pre-training each method.
}
\label{table:detection}
\end{table}

%% file: tables/ablation_sk.tex
\begin{table}[t]
\small
\setlength{\tabcolsep}{4pt}
\centering
\begin{tabular}{@{}lccc@{}}
\toprule
    & \multicolumn{3}{c}{nuScenes} \\
\cmidrule{2-4}
    & $\bevstride=20$\,cm & $\bevstride = 30$\,cm & $\bevstride = 40$\,cm \\
\midrule
    $\Delta_{\rm time} = 0.5$\,s & 58.1 & 57.4 & 56.6
\\
    $\Delta_{\rm time} = 1.0$\,s   & 58.2 & \textbf{58.7} & 57.2
\\
    $\Delta_{\rm time} = 1.5$\,s & 58.5 & 57.1 & 56.7 
\\
    $\Delta_{\rm time} = 2.0$\,s   & 58.3 & 58.3 & 57.6 
\\
\bottomrule
\end{tabular}
\caption{
\textbf{Sensitivity to $\Delta_{\rm time}$}, the time difference between the capture of $\pcone$ and $\pctwo$, and $\bevstride$, the size of the BEV cells for pooling. The backbone is pre-trained on 600 scenes extracted from the original training set of nuScenes, and fine-tuned on 60 scenes. We report the mIoU\% on the spared 100 scenes from the original training set of nuScenes.
}
\label{table:ablation_nuscenes}
\end{table}

\begin{table}[t]
\small
\setlength{\tabcolsep}{4pt}
\centering
\begin{tabular}{@{}lccc@{}}
\toprule
    & \multicolumn{3}{c}{SemanticKITTI} \\
\cmidrule{2-4}
    & $\bevstride=10$ cm & $\bevstride = 20$ cm & $\bevstride = 40$ cm \\
\midrule
    $\Delta_{\rm time} = 0.5$\,s & -    & 60.3 & 59.9 
\\
    $\Delta_{\rm time} = 0.7$\,s & 58.8 & \textbf{61.3} & 60.4 
\\
    $\Delta_{\rm time} = 0.9$\,s & -    & 60.7 & - 
\\
\bottomrule
\end{tabular}
\caption{
\textbf{Sensitivity to $\Delta_{\rm time}$}, the time difference between the capture of $\pcone$ and $\pctwo$, and $\bevstride$, the size of the BEV cells for pooling. The backbone is pre-trained on the full training set of SemanticKITTI, and fine-tuned on 10\% of the training set. We report the mIoU\% on the validation set of SemanticKITTI.
}
\label{table:ablation_sk}
\end{table}

%% file: tables/ablation_dx.tex
\begin{table}[!t]
\small
\setlength{\tabcolsep}{9pt}
\centering
\begin{tabular}{@{}lllllll@{}}
\toprule
\multicolumn{3}{c}{$\Delta_{\rm time}$} & \multicolumn{4}{c}{$\Delta_{\rm dist}$}\\
\cmidrule(lr){1-3}
\cmidrule(lr){4-7}
    $0.5$\,s & $0.7$\,s & $0.9$\,s & $5$\,m & $10$\,m & $15$\,m & $20$\,m\\
\midrule
   60.3 & \textbf{61.3} & 60.7 & 60.0 & 60.5 & 60.6 & 59.6
\\
\bottomrule
\end{tabular}
\caption{
\textbf{Effect on the performance when selecting the point clouds $\pcone$ and $\pctwo$ acquired $\Delta_{\rm time}$ seconds apart or $\Delta_{\rm dist}$ meters apart.} The results are obtained by pre-training, with $\bevstride=20$\,cm, and fine-tuning on SemanticKITTI. We report the mIoU\% on the validation set of SemanticKITTI.
}
\label{table:ablation_dx}
\end{table}

%% file: tables/ablation_registration.tex
\begin{table}[t]
\small
\setlength{\tabcolsep}{4pt}
\centering
\begin{tabular}{@{}ccc@{}}
\toprule
  \it 2D Bi. & \it 2D NN & \it 3D \\
\midrule
  \textbf{61.3} & 59.5 & 60.0
\\
\bottomrule
\end{tabular}
\caption{\textbf{Effect of the registration method.} We analyze the impact of three registration methods to align the representation $\bevone$ and $\bevtwo$. \emph{2D Bi.} and \emph{2D NN} are aligning the representations after projection in BEV, the first using a bilinear interpolation and the second using nearest neighbor interpolation. \emph{3D} is aligning the representation before projection in BEV. The backbones are pre-trained on SemanticKITTI and fine-tuned on the same dataset using $10$\% of annotated scans with $\Delta_{\rm time}=0.7$ s and $\bevstride=20$ cm. We report the mIoU\% on the validation set of SemanticKITTI.}
\label{table:ablation_registration}
\end{table}

%% file: sections/conclusion.tex
\section{Conclusion}

We presented \ours, a simple yet effective contrastive self-supervised method to pre-train 3D backbones for automotive Lidar point clouds. It conveys the surprising observation that contrasting representations of cells on a regular BEV grid performs on par with the most sophisticated contrastive methods, that resort to unsupervised clusterization algorithms. With our method, we are able to reach SOTA results on self-supervised semantic segmentation on nuScenes and SemanticKITTI, while demonstrating the competitiveness of our method on object detection on KITTI~\cite{kitti3d} with a different backbone.

% DO NOT INCLUDE FOR SUBMISSION
\paragraph{Acknowledgements:}
This work was performed using HPC resources from GENCI–IDRIS (Grants 2023-AD011013765).
This work was supported in part by the French Agence Nationale de la Recherche (ANR) grant MultiTrans (ANR21-CE23-0032)

%% file: sections/supmat.tex
\clearpage
\setcounter{page}{1}
\maketitlesupplementary
\appendix

\section{Alignment of the BEV representations}

\subsection{Registration}

Our loss is constructed by aligning the BEV representations $\bevone$ and $\bevtwo$. 

One possibility to align these representations is to register $\pctwo$ onto $\pcone$ after computing $f_\theta(\pctwo)$ but before projection in BEV, i.e., computing $g(f_\theta(\pctwo), r(\pctwo))$ (see section 4.4.3). For each point $p'_i$ with coordinates $(x'_i, y'_i, z'_i)$ in $\pctwo$, the corresponding transformed coordinates $r(\pctwo)_i = (X'_i, Y'_i, Z'_i)$ satisfies    
\begin{align}
\begin{pmatrix}
X'_i\\  
Y'_i \\
Z'_i \\
\end{pmatrix} 
=& \begin{pmatrix}
R_{11} & R_{12} & R_{13}\\ 
R_{21} & R_{22} & R_{23}\\ 
R_{31} & R_{32} & R_{33}
\end{pmatrix}
\begin{pmatrix}
x'_i \\  
y'_i \\
z'_i \\
\end{pmatrix} + 
\begin{pmatrix}
t_1 \\  
t_2 \\
t_3 \\
\end{pmatrix} 
\\
=& \begin{pmatrix}
R_{11} x'_i + R_{12} y'_i + R_{13} z'_i + t_1\\  
R_{21} x'_i + R_{22} y'_i + R_{23} z'_i + t_2\\
R_{31} x'_i + R_{32} y'_i + R_{33} z'_i + t_3\\
\end{pmatrix},
\label{eq:3d_reg}
\end{align}
where we used the rotation matrix $R$ and translation vector $t$ definted in Section 3.1. The function $g(\cdot, \cdot)$ then averages the representation of the points whose coordinates $(X'_i, Y'_i)$ falls in the same BEV cell. Unfortunately, our experiments (see Section 4.4.3) show that this approach leads to suboptimal results.

The approach we advocate for consists in projecting and pooling the representations before alignment, i.e., computing $g(f_\theta(\pctwo), \pctwo)$ (where the representations of the points whose coordinates $(x'_i, y'_i)$ fall in the same BEV cell are averaged) and then aligning the representation by bilinear interpolation. As the projection in BEV makes us loose all access about the actual value of $z'_i$, we approximate the value of $(X'_i, Y'_i)$ in \eqref{eq:3d_reg} by $(X''_i, Y''_i)$ where
\begin{align}
\begin{pmatrix}
X''_i\\  
Y''_i \\
\end{pmatrix} 
= \begin{pmatrix}
R_{11} x'_i + R_{12} y'_i + t_1\\  
R_{21} x'_i + R_{22} y'_i + t_2\\
\end{pmatrix}.
\label{eq:3d_reg_approx}
\end{align}
This approximation holds when $\vert R_{13} z'_i \vert \ll \vert R_{11} x'_i + R_{12} y'_i + t_1 \vert $ and $\vert R_{23} z'_i \vert \ll \vert R_{21} x'_i + R_{22} y'_i + t_2 \vert$, e.g., when there are only small rotations around the $x$ and $y$-axis (often observed in practice on our datasets) or for all points whose coordinates satisfy $z'_i \approx 0$.
Furthermore, this formulation allows easy adaptation to backbones with outputs in the BEV plane, such as those of Second~\cite{second} and PVRCNN~\cite{pvrcnn}.

\subsection{Bilinear interpolation details.} 

Empty cells in the BEV representation $\bevtwo$ are zero-valued, and so are padding cells at the border of the grid.
The interpolation to obtain $\tilde{\mathcal{B}}'$ from $\bevtwo$ being bilinear, the features in each cell after transformation will be a weighted sum of features before the transformation, some of which might be zero vectors. To avoid the influence of the varying number of empty cells around cell filled cells during interpolation, one might want to rescale (cell-wise) the interpolated BEV representation $\tilde{\mathcal{B}}'$  to take into account the number of empty cells involed in the interpolation. This cell-wise rescaling, which would increase the complexity of implementation, is fortunately not necessary because the interpolated feature in each cell of $\tilde{\mathcal{B}}'$ are $\ell_2$-normalized immediately after for computing the loss.

\section{Detailed results and visualizations}

We present in Table~\ref{tab:details_sk} and Table~\ref{tab:details_ns} the per clas IoUs obtained on SemanticKITTI and nuScenes, respectively, for \ours and others when available. On SemanticKITTI, \ours performs better for most classes. 

For completeness, we report in Table~\ref{tab:details_ns}, Table~\ref{tab:details_sk}, Table~\ref{tab:details_sp} the mIoU of each of the five individual runs performed to obtain the main results in the paper. 

Figure~\ref{fig:visu_semseg_ablation} illustrates the better segmentation quality on the class other vehicles of \ours compared to ALSO~\cite{also} and TARL~\cite{TARL}.

Finally, we present in Table~\ref{tab:adetails_kitti3d} all object detection metrics automatically computed in OpenPCDet on KITTI3D for our method, ALSO and no pretraining. The metric use the Average Precision with 40 recall positions where a correct prediction is a box with an overlap of at least 70\% for a car and 50\% for a pedestrian or a cyclist. The precisions are provided in the $2$D image plane, BEV horizontal plane, as well as using the $3$D coordinates. The ``orientation similarity'' corresponds to the Average Orientation Similarity for the correct prediction. 
It is defined as the average over all correct predictions of 
$({1 + \cos (\delta)})/{2}$ 
where $\delta$ is the difference between the predicted and ground truth orientations for any matching boxes.

\input{tables/per_class_ns}
\input{tables/per_class_sk}
\input{tables/details_ns}
\input{tables/details_sk}
\input{tables/details_sp}
\input{tables/details_kitti3d}

\input{figures/visus_semseg_ablation}

%% file: tables/per_class_ns.tex
\begin{table*}[ht!]
\small
\centering
\newcommand*\rotext{\multicolumn{1}{R{70}{1em}}}
\setlength{\tabcolsep}{4pt}
\begin{tabular}{@{}l|llllllllllllllllll}
\toprule
    Method
    & \rotext{\bf mIoU}
    & \rotext{barrier}
    & \rotext{bicycle}
    & \rotext{bus}
    & \rotext{car}
    & \rotext{const. veh.}
    & \rotext{motorcycle}
    & \rotext{pedestrian}
    & \rotext{traffic cone}
    & \rotext{trailer}
    & \rotext{truck}
    & \rotext{driv. surf.}
    & \rotext{other flat}
    & \rotext{sidewalk}
    & \rotext{terrain}
    & \rotext{manmade}
    & \rotext{vegetation}
    \\
\midrule
    \ours
    & 37.9
    & 0.0
    & 1.3
    & 32.6
    & 74.3
    & 1.1
    & 0.9
    & 41.3
    & 8.1
    & 24.1
    & 40.9
    & 89.8
    & 36.2
    & 44.0
    & 52.1
    & 79.9
    & 79.7
    \\
\bottomrule
\end{tabular}
\caption{Per-class IoU on nuScenes using $1\%$ of the annotated scans for fine-tuning. Results are averaged over $5$ different fine-tunings.
Only \ours per-class results can be given as previous methods did not detail those information.}
\label{tab:per_class_ns}
\end{table*}

%% file: tables/per_class_sk.tex
\begin{table*}[ht!]
\small
\centering
\newcommand*\rotext{\multicolumn{1}{R{70}{1em}}}
\setlength{\tabcolsep}{3.46pt}
\begin{tabular}{@{}l|lllllllllllllllllllll}
\toprule
    Method
    & \rotext{\bf mIoU}
    & \rotext{car}
    & \rotext{bicycle}
    & \rotext{motorcycle}
    & \rotext{truck}
    & \rotext{other-veh.}
    & \rotext{person}
    & \rotext{bicyclist}
    & \rotext{motorcycli.}
    & \rotext{road}
    & \rotext{parking}
    & \rotext{sidewalk}
    & \rotext{oth. ground}
    & \rotext{building}
    & \rotext{fence}
    & \rotext{vegetation}
    & \rotext{trunk}
    & \rotext{terrain}
    & \rotext{pole}
    & \rotext{traffic sign}
    \\
\midrule
    STSSL~\cite{stssl} & 49.4 & 92.8 & 8.3 & 27.2 & 32.0 & 25.3 & 48.3 & 57.7 & 0.0 & 90.6 & 31.2 & 73.9 & 0.7 & 87.7 & 43.3 & 86.6 & 57.1 & 72.4 & 57.9 & 46.2
    \\
    TARL~\cite{TARL}
    & 52.5
    & 93.4
    & 16.6
    & 45.1
    & 43.5
    & 21.7
    & 49.4
    & 61.8
    & 0.1
    & 91.0
    & 33.4
    & 75.1
    & 0.5
    & 89.0
    & 43.8
    & 87.5
    & 61.7
    & 74.0
    & 59.9
    & 50.3\\
    \ours 
    & 53.8
    & 93.8
    & 15.6
    & 39.0
    & 50.1
    & 28.9
    & 52.5
    & 61.8
    & 0.2
    & 92.0
    & 40.1
    & 76.1
    & 0.5
    & 90.4
    & 49.6
    & 87.7
    & 61.1
    & 73.7
    & 60.6
    & 48.5\\
\bottomrule
\end{tabular}
\caption{Per-class IoU\% on SemanticKITTI using $1\%$ of the annotated scans for fine-tuning. Results are averaged over $5$ different fine-tunings.}
\label{tab:per_class_sk}
\end{table*}

%% file: tables/details_ns.tex
\begin{table*}[ht!]

\centering
\small

\begin{tabular}{cl|ccccc|c}
\toprule
\% & Method          & \multicolumn{5}{c|}{Runs} & Average and std \\
\midrule
0.1\% & No pre-training     & 21.88 & 21.21 & 22.05 & 21.08 & 21.96 & 21.64 $\pm$0.45\\
& ALSO~\cite{also} & 26.62 & 26.86 & 25.99 & 25.59 & 26.08 & 26.23 $\pm$0.51\\
&\ours & 27.25&	26.47	&26.21&	26.06	&26.76&	26.55 $\pm$0.47 \\

\midrule\midrule
1\% & No pre-training		& 34.86	& 35.09	& 34.72	& 34.72	& 35.55	& 34.99 $\pm$0.35\\
& ALSO~\cite{also}& 37.42	& 37.52	& 37.15	& 37.11	& 37.94	& 37.43 $\pm$0.34\\
& \ours &37.66	&38.28&	38.16	&37.41	&38.01	&37.90 $\pm$0.36\\

\midrule\midrule
10\% & No pre-training		& 57.62	& 57.66	& 57.31	& 56.70	& 57.19	& 57.30 $\pm$0.39\\
& ALSO~\cite{also}& 58.63	& 58.62	& 59.11	& 59.28	& 59.35	& 59.00 $\pm$0.35\\
&\ours &58.29	&58.66&	58.74&	59.77&	59.58	&59.01 $\pm$0.64\\

\midrule\midrule
50\% & No pre-training		& 68.80	& 68.90	& 68.94	& 69.31	& 69.01	& 68.99 $\pm$0.19\\
& ALSO~\cite{also}& 69.69	& 69.58	& 69.93	& 69.66	& 70.17	& 69.81 $\pm$0.24\\
&\ours &70.14	&70.47	&70.80&	70.59	&70.59&	70.52 $\pm$0.24\\

\midrule\midrule
100 \% & No pre-training     & 71.21	& 71.35	& 71.20	& 70.93	& 71.32	& 71.20 $\pm$0.17\\
& ALSO~\cite{also}& 71.95	& 71.92	& 71.60	& 71.88	& 71.39	& 71.75 $\pm$0.24\\
&\ours & 72.17&	72.18&	71.97	&72.11	&72.34	&72.15 $\pm$0.13\\
\bottomrule
\end{tabular}

\caption{Details of individual run on nuScenes semantic segmentation. We report the mIoU\% on the official validation set for each of the individual runs averaged in the main paper.}
\label{tab:details_ns}
\end{table*}

%% file: tables/details_sk.tex
\begin{table*}[ht!]

\centering
\small

\begin{tabular}{cl|ccccc|c}
\toprule
\% & Method          & \multicolumn{5}{c|}{Runs} & Average and std \\
\midrule
0.1\% & No pre-training & 30.22 & 29.99 & 29.74 & 30.15 & 29.77 & 29.97 $\pm$0.22\\
& STSSL~\cite{stssl} & 31.37&32.62&32.07&32.03&32.01&32.02 $\pm$0.44\\
& ALSO~\cite{also} & 34.97 & 34.83 & 34.81 & 35.10 & 35.11 & 34.96 $\pm$0.14\\
& TARL~\cite{TARL} & 38.01	&38.25	&37.89&	37.87	&37.28	&37.86 $\pm$0.36\\
&\ours & 39.08	&40.28	&38.42&	40.53	&40.31	&39.72 $\pm$0.92 \\

\midrule\midrule
1\% & No pre-training & 45.1 & 46.32	& 46.59	& 46.68	& 46.49 & 46.24	$\pm$0.65\\
& STSSL~\cite{stssl} & 50.03&	48.90	&51.00	&48.22&	49.00&	49.43 $\pm$1.09\\
& ALSO~\cite{also} & 50.04	& 50.28	& 49.43	& 50.41	& 50.02	& 50.04 $\pm$0.38\\
& TARL~\cite{TARL} & 52.09&	52.41&	52.35&	53.43&	52.32&	52.52 $\pm$0.52\\
& \ours&53.59	&54.28	&53.86&	54.92	&52.35	&53.80 $\pm$0.95 \\

\midrule\midrule
10\% & No pre-training & 57.04 & 58.74 & 57.71 & 56.27 & 58.01 & 57.55 $\pm$0.94\\
& STSSL~\cite{stssl} & 60.09&	59.19&	60.65&	60.29&	59.69	&59.98 $\pm$0.56\\
& ALSO~\cite{also} & 60.41 & 60.45 & 60.47 & 60.54 & 60.43 & 60.46 $\pm$0.05\\
& TARL~\cite{TARL} & 61.04	&60.92	&61.64	&60.97&	61.19&	61.15 $\pm$0.29\\
&\ours & 61.82&	60.74&	61.74&	61.28	&61.31&	61.38 $\pm$0.43 \\

\midrule\midrule
50\% & No pre-training & 61.48 & 62.33 & 61.88 & 61.80 & 61.31 & 61.76 $\pm$0.39\\
& STSSL~\cite{stssl} & 63.50&	63.33&	61.91	&62.54	&63.29	&62.91 $\pm$0.67\\
& ALSO~\cite{also} & 63.09 & 63.43 & 62.99 & 63.28 & 64.15 & 63.39 $\pm$0.46\\
& TARL~\cite{TARL} & 63.23	&63.66&	63.43	&63.25	&63.27&	63.37 $\pm$0.18\\
&\ours&62.43&63.27	&64.07&	63.77&	63.65&	63.44 $\pm$0.63\\

\midrule\midrule
100 \% & No pre-training     & 62.49 & 62.35 & 62.98 & 62.50 & 63.06	& 62.68 $\pm$0.32\\
& STSSL~\cite{stssl}&63.27&	63.50	&63.33	&62.74	&63.66	&63.30 $\pm$0.35\\
& ALSO~\cite{also} & 64.29 & 63.75 & 63.75 & 63.34 & 63.07 & 63.64 $\pm$0.46\\
& TARL~\cite{TARL}&63.83	&63.71&	63.14	&63.90&	63.92&	63.70 $\pm$0.32\\
&\ours & 63.94	&63.84&	64.05	&64.74	&63.73	&64.06 $\pm$0.40\\
\bottomrule
\end{tabular}

\caption{Details of individual run on SemanticKITTI semantic segmentation. We report the mIoU\% on the official validation set for each of the individual runs averaged in the main paper.}
\label{tab:details_sk}
\end{table*}

%% file: tables/details_sp.tex
\begin{table*}[ht!]

\centering
\small

\begin{tabular}{cl|ccccc|c}
\toprule
\% & Method             & \multicolumn{5}{c|}{Runs} & Average and std \\
\midrule
0.1\% & No pre-training & 37.23 & 37.53 & 36.37 & 36.72 & 36.59 & 36.89 $\pm$0.48 \\
& STSSL~\cite{stssl}    & 41.04&39.97&40.05&39.78&40.33&40.23 $\pm$0.49\\
& ALSO~\cite{also}      & 40.04	& 41.23 & 41.29 & 40.88 & 39.83 & 40.65 $\pm$0.68\\
& TARL~\cite{TARL}      & 45.38&46.33&45.29&45.40&46.61&45.80 $\pm$0.62\\
&\ours                  & 43.06&42.92&42.03&43.25&43.04&42.86 $\pm$0.48\\

\midrule\midrule
1\% & No pre-training   & 46.99 & 46.23 & 46.09 & 46.33 & 46.47 & 46.42	$\pm$0.35\\
& STSSL~\cite{stssl}    & 48.86&48.58&49.65&49.55&49.57&49.24 $\pm$0.49\\
& ALSO~\cite{also}      & 50.55 & 48.85 & 49.02 & 49.86 & 49.51 & 49.56	$\pm$0.68\\
& TARL~\cite{TARL}      & 52.24&52.34&52.11&52.85&51.45&52.20 $\pm$0.50\\
& \ours     	        & 50.91&52.13&52.88&52.03&51.48&51.89 $\pm$0.74\\

\midrule\midrule
10\% & No pre-training  & 54.29 & 53.96 & 54.53 & 54.95 & 54.60 & 54.47 $\pm$0.37\\
& STSSL~\cite{stssl}    & 55.03&55.10&55.96&55.84&55.43&55.47 $\pm$0.42\\
& ALSO~\cite{also}      & 56.16 & 56.19 & 55.43 & 55.15 & 56.14 & 55.81 $\pm$0.49\\
& TARL~\cite{TARL}      & 56.29&55.52&56.02&56.12&55.95&55.98 $\pm$0.29\\
&\ours                  & 55.10&55.92&55.50&55.12&55.60&55.45 $\pm$0.35\\

\midrule\midrule
50\% & No pre-training  & 55.48 & 55.19 & 55.40 & 55.27 & 55.04 & 55.28 $\pm$0.17\\
& STSSL~\cite{stssl}      & 55.62&57.96&56.68&57.01&56.72&56.80 $\pm$0.84\\
& ALSO~\cite{also}      & 57.07 & 55.56 & 56.71 & 56.13 & 56.30 & 56.35 $\pm$0.58\\
& TARL~\cite{TARL}      & 55.97&56.86&56.98&56.86&57.09&56.75 $\pm$0.45\\
& \ours     	        & 56.65&55.84&56.96&57.21&57.01&56.73 $\pm$0.54\\

\midrule\midrule
100 \% & No pre-training& 54.52 & 55.52 & 55.52 & 55.10 & 54.83 & 55.10 $\pm$0.44\\
& STSSL~\cite{stssl}      & 54.71&56.34&56.56&57.48&56.09&56.24 $\pm$1.00\\
& ALSO~\cite{also}      & 56.88 & 58.23 & 55.45 & 56.01 & 56.88 & 56.69 $\pm$1.05\\
& TARL~\cite{TARL}      & 56.70&56.42&57.92&57.01&56.94&57.00 $\pm$0.56\\
& \ours     	        & 56.31&56.50&56.83&56.21&55.92&56.35 $\pm$0.34\\

\bottomrule
\end{tabular}

\caption{Details of individual run on SemanticPOSS semantic segmentation. We report the mIoU\% on the official validation set with a pre-training on SemanticKITTI for each of the individual runs averaged in the main paper.}
\label{tab:details_sp}
\end{table*}

%% file: tables/details_kitti3d.tex
\begin{table*}[ht!]
\small
\centering

\setlength{\tabcolsep}{4pt}

\begin{tabular}{lll|ccc|ccc|ccc}
\toprule
Backbone & Metric & Method & \multicolumn{3}{c|}{Car} & \multicolumn{3}{c|}{Pedestrian} & \multicolumn{3}{c}{Cyclist} \\
& & & Easy & Mod. & Hard &Easy& Mod. & Hard & Easy& Mod. & Hard\\
\midrule
SECOND & 2D object detection 
& Scratch$^\dagger$      & 95.84 & 94.49 & 92.00 & 68.27 & 64.69 & 61.15 & 91.02 & 78.88 & 76.00 \\
\greyrule{\cmidrule{3-12}}
&& ALSO~\cite{also}             & 97.48 & 94.70 & 93.56 & 72.39 & 68.69 & 65.86 & 91.34 & 81.64 & 78.46 \\
& & \ours & 97.45&	94.53&	93.42& 71.93&	67.33&	63.84& 93.61&	80.85&	77.30\\
\cmidrule{2-12}
&Bird's eye view 
& Scratch$^\dagger$       & 93.76 & 89.82 & 87.65 & 59.74 & 54.85 & 50.56 & 87.19 & 70.96 & 68.00 \\
\greyrule{\cmidrule{3-12}}
&& ALSO~\cite{also}             & 94.64 & 90.77 & 88.24 & 64.32 & 59.13 & 55.25 & 86.92 & 74.58 & 70.17 \\
& & \ours & 92.33&	89.79&	87.47& 61.45&	56.46&	52.74& 90.64&	72.86&	68.35\\
\cmidrule{2-12}
&3D object detection
& Scratch$^\dagger$       & 90.20 & 81.50 & 78.61 & 53.89 & 48.82 & 44.56 & 82.59 & 65.72 & 62.99 \\
\greyrule{\cmidrule{3-12}}
&& ALSO~\cite{also}             & 90.21 & 81.78 & 78.97 & 59.56 & 54.24 & 50.27 & 81.12 & 68.19 & 64.10 \\
& & \ours & 90.18&	81.44&	78.60&56.60&	51.95&	47.64& 87.35&	69.28&	65.06\\
\cmidrule{2-12}
&Orientation similarity
& Scratch$^\dagger$       & 95.83 & 94.35 & 91.79 & 63.70 & 59.52 & 55.85 & 90.86 & 78.44 & 75.52 \\
\greyrule{\cmidrule{3-12}}
&& ALSO~\cite{also}             & 97.45 & 94.54 & 93.30 & 67.67 & 63.33 & 60.28 & 91.01 & 80.75 & 77.49 \\
& & \ours & 97.42&94.42&93.22&66.86&61.80&58.16&93.36&79.67&76.15\\
\midrule\midrule
PV-RCNN & 2D object detection
& Scratch$^\dagger$        & 97.86 & 94.39 & 93.92 & 73.84 & 68.68 & 65.53 & 94.34 & 81.89 & 77.36 \\
\greyrule{\cmidrule{3-12}}
&& ALSO~\cite{also}             & 96.12 & 94.45 & 93.99 & 73.70 & 68.70 & 65.31 & 94.61 & 81.86 & 78.67\\
& & \ours & 97.55&	94.40&	94.04&72.65&	67.55&	64.77&96.42&	82.74&	79.70\\
\cmidrule{2-12}
&Bird's eye view
& Scratch$^\dagger$       & 94.65 & 90.61 & 88.56 & 68.28 & 60.62 & 55.95 & 92.52 & 75.03 & 70.40 \\
\greyrule{\cmidrule{3-12}}
&& ALSO~\cite{also}             & 93.10 & 90.64 & 88.53 & 68.72 & 60.92 & 56.96 & 93.11 & 76.74 & 73.06\\
& & \ours & 92.79&	90.61&	88.49&68.01&	61.40&	57.49&93.70&	74.96&	71.22\\
\cmidrule{2-12}
& 3D object detection
& Scratch$^\dagger$        & 91.74 & 84.60 & 82.29 & 65.51 & 57.49 & 52.71 & 91.37 & 71.51 & 66.98 \\
\greyrule{\cmidrule{3-12}}
&& ALSO~\cite{also}             & 92.31 & 84.86 & 82.61 & 65.60 & 57.76 & 52.96 & 91.70 & 74.98 & 70.67\\
& & \ours & 91.48&	84.47&	82.37&64.85&	58.21&	53.36&90.90&	71.80&	68.02\\
\cmidrule{2-12}
& Orientation similarity
& Scratch$^\dagger$        & 97.84 & 94.25 & 93.70 & 69.73 & 63.89 & 60.31 & 94.20 & 81.00 & 76.47 \\
\greyrule{\cmidrule{3-12}}
&& ALSO~\cite{also}             & 96.09 & 94.29 & 93.76 & 67.66 & 62.74 & 59.37 & 94.23 & 81.07 & 77.86\\
& & \ours & 97.52&	94.27&	93.83&67.26&	61.99&	58.98&96.16&	82.14&	79.11\\
\bottomrule
\end{tabular}

\caption{Details of object detection metrics on KITTI3D with a pre-training on nuScenes, for all difficulties. Details of the metrics are provided above. The results presented in the main paper use the ``3D object detection'' metric.}
\label{tab:adetails_kitti3d}
\end{table*}

%% file: figures/visus_semseg_ablation.tex
\begin{figure*}
    \centering
    \setlength{\tabcolsep}{3pt}
    \begin{tabular}{@{}cccc@{}}
        \includegraphics[width=0.24\linewidth]{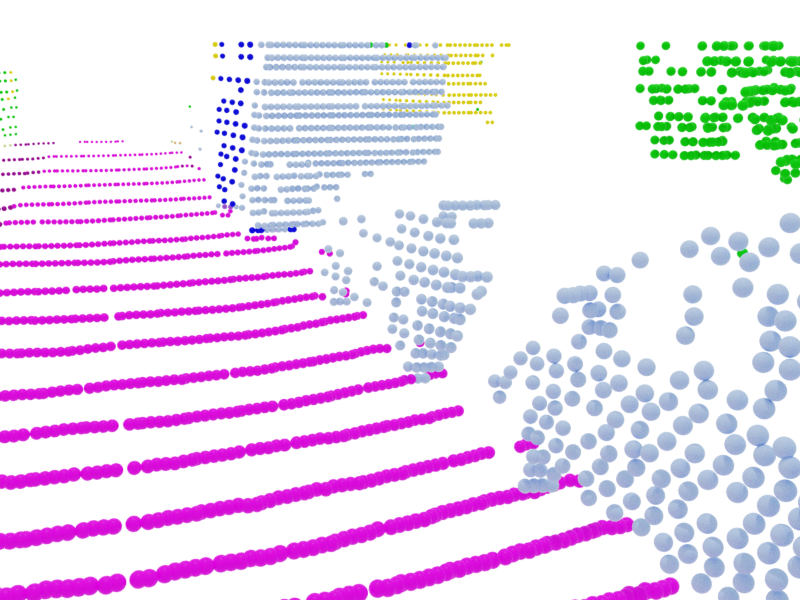} &
        \includegraphics[width=0.24\linewidth]{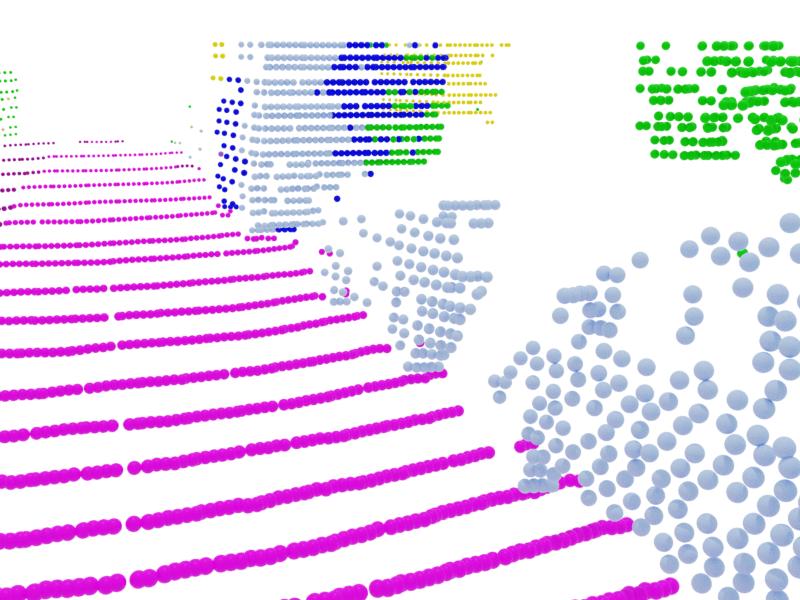} &
        \includegraphics[width=0.24\linewidth]{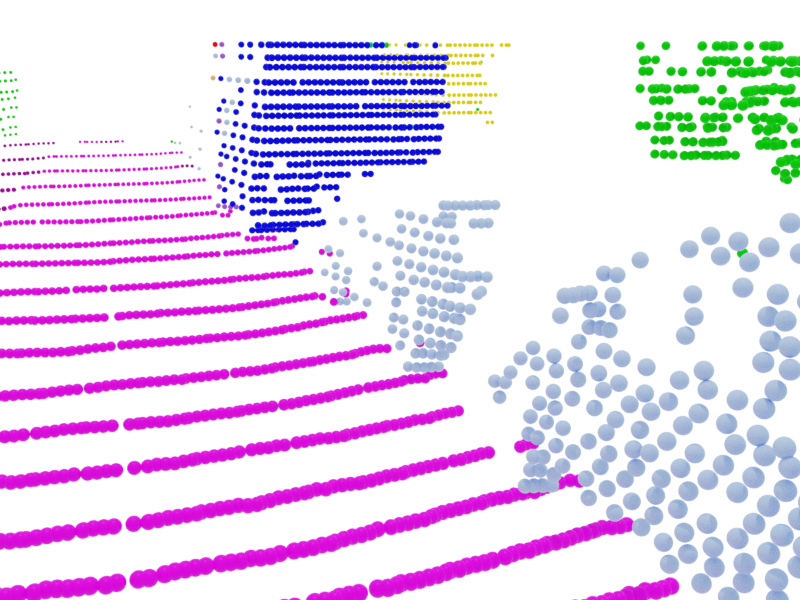} &
        \includegraphics[width=0.24\linewidth]{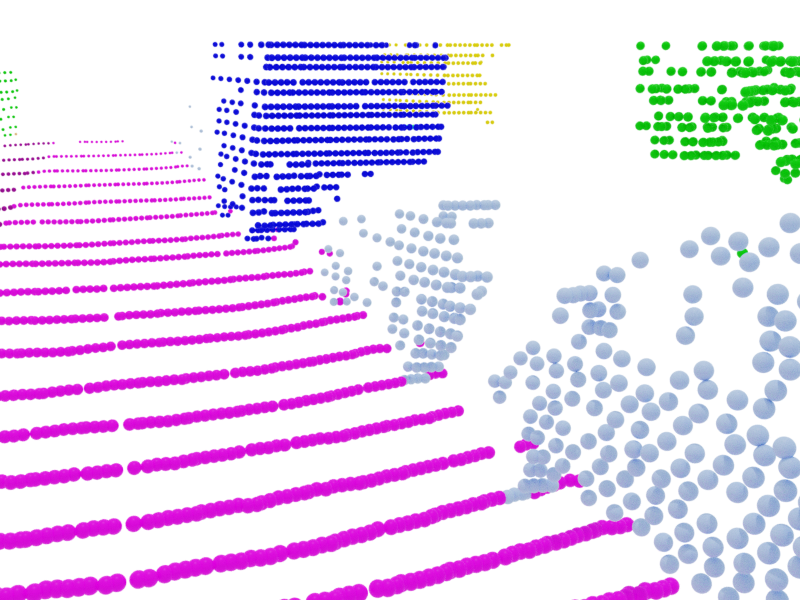} \\
        ALSO & TARL & \ours & Ground truth\\
    \end{tabular}
    \caption{\textbf{Semantic segmentation visualizations on SemanticKITTI} after pre-training on the full training dataset and fine-tuning using 1\% of the annotated data.
    \colorbox{skcar}{\color{white}car\vphantom{$X_p$}},
    \colorbox{skotherv}{\color{white}other veh.\vphantom{$X_p$}},
    \colorbox{skdrivable}{\color{white}drivable surf.\vphantom{$X_p$}},
    \colorbox{skvegetation}{\color{white}vegetation\vphantom{$X_p$}}
}
    \label{fig:visu_semseg_ablation}
\end{figure*}